# Generative Adversarial Perturbations

Omid Poursaeed[1,2] Isay Katsman[1] Bicheng Gao[3,1] Serge Belongie[1,2]
[1]Cornell University [2]Cornell Tech [3]Shanghai Jiao Tong University
{op63,isk22,bg455,sjb344}@cornell.edu

## Abstract

*In this paper, we propose novel generative models for creating adversarial examples, slightly perturbed images resembling natural images but maliciously crafted to fool pre-trained models. We present trainable deep neural networks for transforming images to adversarial perturbations. Our proposed models can produce image-agnostic and image-dependent perturbations for targeted and non-targeted attacks. We also demonstrate that similar architectures can achieve impressive results in fooling both classification and semantic segmentation models, obviating the need for hand-crafting attack methods for each task. Using extensive experiments on challenging high-resolution datasets such as ImageNet and Cityscapes, we show that our perturbations achieve high fooling rates with small perturbation norms. Moreover, our attacks are considerably faster than current iterative methods at inference time*[1].

## 1. Introduction

In spite of their impressive performance on challenging tasks in computer vision such as image classification [25, 49, 51, 52, 20] and semantic segmentation [30, 5, 9, 59, 60], deep neural networks are shown to be highly vulnerable to adversarial examples, i.e. carefully crafted samples looking similar to natural images but designed to mislead a pre-trained model. This phenomenon was first studied in [53], and may hinder the applications of deep networks on visual tasks, or pose security concerns.

Two types of adversarial perturbations can be considered: Universal and Image-dependent. Image-dependent perturbations can vary for different images in the dataset. To generate these perturbations, we require a function which takes a natural image, and outputs an adversarial image. We approximate this function with a deep neural network. Universal perturbations are fixed perturbations which when added to natural images can significantly degrade the accuracy of the pre-trianed network. In this case, we seek a perturbation $\mathcal{U}$ with small magnitude such that for most natural images $x$, $x + \mathcal{U}$ can fool the pre-trained model. Unlike the iterative approaches proposed in the literature, we consider trainable networks for learning the universal perturbation.

From another viewpoint, adversarial attacks can be categorized as targeted and non-targeted. In targeted adversarial attacks, we seek adversarial images that can change the prediction of a model to a specific target label. In non-targeted attacks we want to generate adversarial examples for which the model's prediction is any label other than the ground-truth label. Considering all the possible combinations, we can have four types of adversarial examples: targeted universal, non-targeted universal, targeted image-dependent and non-targeted image-dependent. We elaborate on each of them in the following sections.

Our main contributions can be summarized as follows:

- We present a unifying framework for creating universal and image-dependent perturbations for both classification and semantic segmentation tasks, considering targeted and non-targeted attacks with $L_\infty$ and $L_2$ norms as the metric.
- We improve the state-of-the-art performance in universal perturbations by leveraging generative models in lieu of current iterative methods.
- We are the first to present effective targeted universal perturbations. This is the most challenging task as we are constrained to have a single perturbation pattern and the prediction should match a specific target.
- Our attacks are considerably faster than iterative and optimization-based methods at inference time. We can generate perturbations in the order of milliseconds.

## 2. Related Work

### 2.1. Universal Perturbations

First introduced in [35], universal perturbations are fixed perturbations which after being added to natural images can mislead a pre-trained model for most of the images. The algorithm in [35] iterates over samples in a target set, and gradually builds the universal perturbation by aggregating image-dependent perturbations and normalizing the result. [38] presents a data independent approach for generating image-agnostic perturbations. Its objective is to maximize the product of mean activations at multiple layers of the network when the input is the universal perturbation. While this method obviates the need for training data, the results are not as strong as [35]. A method for generating targeted universal adversarial perturbations for semantic segmentation models is presented in [34]. Their approach is similar to [35] in that they also create the universal perturbation by adding image-dependent perturbations and clipping the result to limit the norm. [36] proposes a quantitative analysis of the robustness of classifiers to universal perturbations based on the geometric properties of decision boundaries. A defense method against universal adversarial perturbations is proposed in [1]. It learns a Perturbation Rectifying Network (PRN) from real and synthetic universal perturbations, without needing to modify the target model.

[1]Code is available at https://github.com/OmidPoursaeed/Generative_Adversarial_Perturbations.

### 2.2. Image-dependent Perturbations

Various approaches have been proposed for creating image-dependent perturbations. Optimization-based methods such as [53] and [8] define a cost function based on the perturbation norm and the model's loss. Then they use gradient ascent in pixel space with optimizers such as L-BFGS or Adam [24] to create the perturbation. While these approaches yield better results than other methods, they are slow at inference time as they need to forward the input to the model several times.

[18] proposes a Fast Gradient Sign Method (FGSM) to generate adversarial examples. It computes the gradient of the loss function with respect to pixels, and moves a single step based on the sign of the gradient. While this method is fast, using only a single direction based on the linear approximation of the loss function often leads to sub-optimal results. Based on this work, [37] presents an iterative algorithm to compute the adversarial perturbation by assuming that the loss function can be linearized around the current data point at each iteration. [26] introduces the *Iterative Least-Likely Class* method, an iterative gradient-based method choosing the least-likely prediction as the desired class. This method is applied to ImageNet in [27]. It also discusses how to effectively include adversarial examples in training to increase model's robustness. [11] proposes a method for directly optimizing performance measures, even when they are combinatorial and non-decomposable. [39] generates images unrecognizable to humans but classified with high confidence as members of a recognizable class. It uses evolutionary algorithms and gradient ascent to fool deep neural networks. Our work bears a resemblance to [6] in that it also considers training a network for generating adversarial examples. However, [6] does not provide a fixed bound on the perturbation magnitude, which might make perturbations detectable at inference time. It is also limited to targeted image-dependent perturbations. [58] extends adversarial examples from the task of image classification to semantic segmentation and object detection. For each image, it applies gradient ascent in an iterative procedure until the number of correctly predicted targets becomes zero or a maximum iteration is reached. Similar to [53] and [8], this method suffers from being slow at inference time. [2] evaluates the robustness of segmentation models against common attacks. [31] suggests that adversarial examples are sensitive to the angle and distance at which the perturbed picture is viewed. [4] presents a method for generating adversarial examples that are robust across various transformations.

Several methods have been proposed for defending against adversarial attacks. While our focus is on efficient attacks, we refer the reader to [33, 57, 19, 47, 32, 50, 48, 54, 3, 13, 44, 16, 55, 45, 42, 56] for recent works on defense.

## 3. Generative Adversarial Perturbations

Consider a classification network $\mathcal{K}$ trained on natural images from $C$ different classes. It assigns a label $\mathcal{K}(x) \in \{1, \ldots, C\}$ to each input image $x$[2]. We assume that images are normalized to $[0,1]$ range. Let $\mathfrak{N} \subset [0,1]^n$ represent the space of natural images[3]. We assume that $\mathcal{K}$ achieves a high accuracy on natural images. Therefore, if we denote the correct class for image $x$ by $c_x$, $\mathcal{K}(x) = c_x$ for most $x \in \mathfrak{N}$. Let $\mathcal{A}_\mathcal{K}$ stand for the space of adversarial examples for the network $\mathcal{K}$. Images in $\mathcal{A}_\mathcal{K}$ must resemble a natural image yet be able to fool the network $\mathcal{K}$. Hence, for each $a \in \mathcal{A}_\mathcal{K}$ there exists $x \in \mathfrak{N}$ such that $d(a,x)$ is small and $\mathcal{K}(a) \neq c_x$, where $d(\cdot,\cdot)$ is a distance metric.

This framework can be easily extended to the task of semantic segmentation in which the correct class for each pixel needs to be determined. In this case, the segmentation network $\mathcal{K}$ assigns a label map $\mathcal{K}(x) = (\mathcal{K}(x_1), \ldots, \mathcal{K}(x_n)) \in \{1, \ldots, C\}^n$ to each image $x = (x_1, \ldots, x_n)$. The ground-truth prediction for image $x$ is $c_x = (c_{x_1}, \ldots, c_{x_n})$, and the set of adversarial examples is $\mathcal{A}_\mathcal{K} = \{a \in [0,1]^n \backslash \mathfrak{N} \,|\, \exists\, x \in \mathfrak{N} : d(a,x) < \epsilon\,, \forall\, i \in \{1, \ldots, n\} : \mathcal{K}(a_i) \neq c_{x_i}\}$, where $\epsilon$ is a fixed threshold[4].

### 3.1. Universal Perturbations

Universal Perturbations were first proposed in the seminal work of Dezfooli *et al.* [35]. The paper proposes an iterative algorithm to generate the universal perturbation. It constructs the universal perturbation by adding image-dependent perturbations obtained from [37] and scaling the result. Unlike the iterative approach of [35], we seek an end-to-end trainable model for generating the universal perturbation. Let us denote the set of universal perturbations for the network $\mathcal{K}$ by $\mathcal{U}_\mathcal{K} = \{\mathcal{U} \in [0,1]^n \mid \text{for most } x \in \mathfrak{N} : x + \mathcal{U} \in \mathcal{A}_\mathcal{K}\}$. We do not want the perturbation to directly depend on any input image from the dataset. We seek a function $f : [0,1]^n \to \mathcal{U}_\mathcal{K}$ which can transform a random pattern to the universal perturbation. By changing the input pattern, we can obtain a diverse set of universal perturbations. In practice, we approximate $f(\cdot)$ with a deep neural network $f_\Theta(\cdot)$ with weights $\Theta$. This setting resembles Generative Adversarial Networks (GANs) [17, 43, 15, 28, 21] in which a random vector is sampled from a latent space, and is transformed to a natural-looking image by a generator. In our case the range of the mapping is $\mathcal{U}_\mathcal{K}$ instead of $\mathfrak{N}$, and the generator is trained with a *fooling* loss instead of the discriminative loss used in GANs. We also tried using a combination of *fooling* and *discriminative* losses; however, it led to sub-optimal results.

[2]Note that $x$ may or may not belong to the space of natural images.

[3]For images of height $h$, width $w$ and $c$ channels: $n = h \times w \times c$.

[4]We can also relax the constraint, and require that for *most* pixels the prediction is different from the ground-truth.

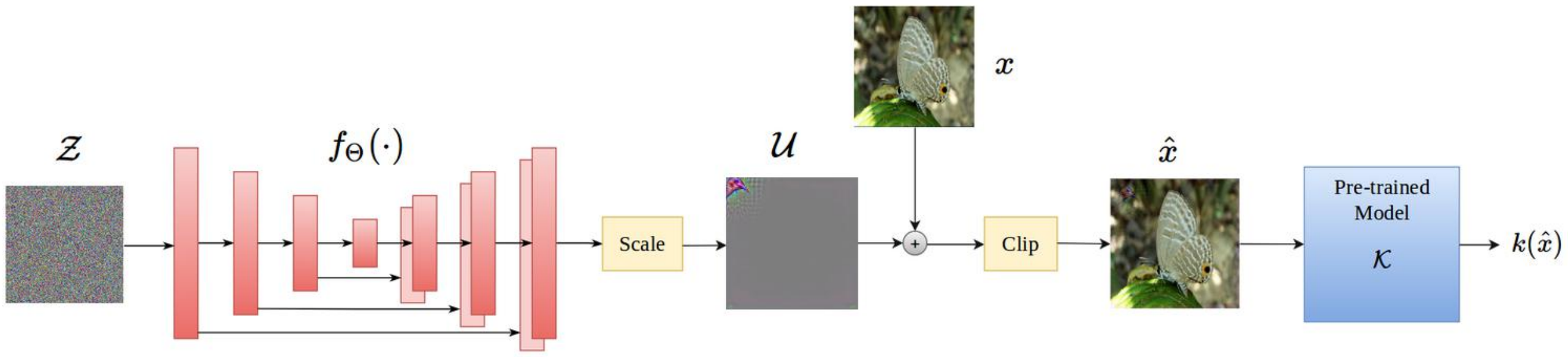

Figure 1: Training architecture for generating universal adversarial perturbations. A fixed pattern, sampled from a uniform distribution, is passed through the generator. The scaled result is the universal perturbation which, when added to natural images, can mislead the pre-trained model. We consider both U-Net (illustrated here) and ResNet Generator architectures.

There are several options for the architecture of the image transformation network $f_\Theta(\cdot)$. We consider two architectures used in recent image-to-image translation networks such as [22] and [61]. The U-Net architecture [46] is an encoder-decoder network with skip connections between the encoder and the decoder. The other architecture is ResNet Generator which was introduced in [23], and is also used in [61] for transforming images from one domain to another. It consists of several downsampling layers, residual blocks and upsampling layers. In most of our experiments, the ResNet Generator outperforms U-Net.

Figure 1 illustrates the architecture for generating universal perturbations. A fixed pattern $\mathcal{Z} \in [0,1]^n$, sampled from a uniform distribution $U[0,1]^n$, is fed to a generator $f_\Theta$ to create the perturbation. The output of the generator $f_\Theta(\mathcal{Z})$ is then scaled to have a fixed norm. More specifically, we multiply it by $\min\left(1, \frac{\epsilon}{\|f_\Theta(\mathcal{Z})\|_p}\right)$ in which $\epsilon$ is the maximum permissible $L_p$ norm. Similar to related works in the literature, we consider $p = 2$ and $p = \infty$ in experiments. The resulting universal perturbation $\mathcal{U}$ is added to natural images to create the perturbed ones. Before feeding the perturbed image to the generator, we clip it to keep it in the valid range of images on which the network is trained. We feed the clipped image $\hat{x}$ to the network $\mathcal{K}$ to obtain the output probabilities $k(\hat{x})$[5]. Let $\mathbb{1}_{c_x}$ denote the one-hot encoding of the ground-truth for image $x$. In semantic segmentation, $c_x \in \{1, \ldots, C\}^n$ is the ground-truth label map, and $k(\hat{x})$ contains the class probabilities for each pixel in $\hat{x}$. For non-targeted attacks we want the prediction $k(\hat{x})$ to be different from $\mathbb{1}_{c_x}$, so we define the loss to be a decreasing function of the cross-entropy $\mathcal{H}(k(\hat{x}), \mathbb{1}_{c_x})$. We found that the following *fooling loss* gives good results in experiments:

$$l_{non-targeted} = l_{fool} = -\log(\mathcal{H}(k(\hat{x}), \mathbb{1}_{c_x})) \quad (1)$$

Alternatively, as proposed by [26] and [27], we can consider the least likely class $k_{ll}(x) = \arg\min k(x)$, and set it as the target for training the model:

$$l_{non-targeted} = l_{fool} = \log(\mathcal{H}(k(\hat{x}), \mathbb{1}_{k_{ll}(x)})) \quad (2)$$

In practice, the losses in equations 1 and 2 lead to competitive results. We also found that for the Inception model, the logit-based loss used in [7, 8] yields optimal results.

For targeted perturbations we consider the cross-entropy with the one-hot encoding of the target:

$$l_{targeted} = l_{fool} = \log(\mathcal{H}(k(\hat{x}), \mathbb{1}_t)) \quad (3)$$

where $t$ represents the target. Note that for the classification task, $t \in \{1, \ldots, C\}$ is the target class while in semantic segmentation, $t \in \{1, \ldots, C\}^n$ is the target label map.

### 3.2. Image-dependent Perturbations

We consider the task of perturbing images as a transformation from the domain of natural images to the domain of adversarial images. In other words, we require a mapping $f : \mathfrak{N} \to \mathcal{A}_\mathcal{K}$ which generates a perturbed image $f(x) \in \mathcal{A}_\mathcal{K}$ for each natural image $x \in \mathfrak{N}$. A desirable function $f(\cdot)$ must result in a low accuracy and a high *fooling ratio*. Accuracy denotes the proportion of samples $x$ for which $\mathcal{K}(f(x)) = c_x$, while fooling ratio represents the ratio of images $x$ for which $\mathcal{K}(f(x)) \neq \mathcal{K}(x)$. Since we assume that the model achieves a high accuracy on natural images, these two metrics are highly correlated.

[5]Note that $\mathcal{K}(\hat{x}) = \arg\max k(\hat{x})$.

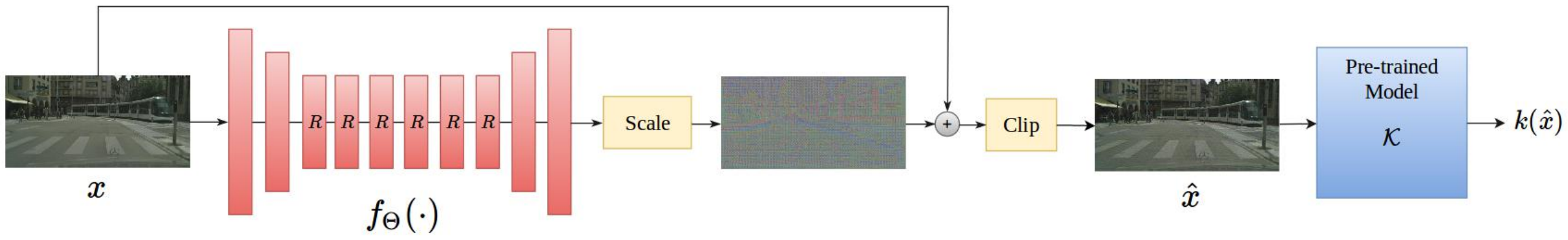

Figure 2: Architecture for generating image-dependent perturbations. The generator outputs a perturbation, which is scaled to satisfy a norm constraint. It is then added to the original image, and clipped to produce the perturbed image. We use the ResNet Generator architecture for most of the image-dependent tasks.

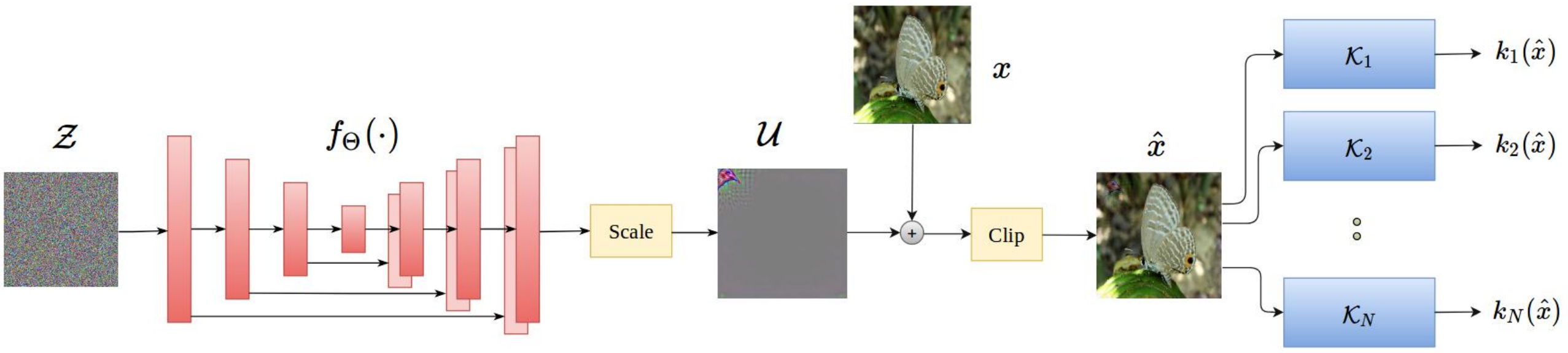

Figure 3: Architecture for training a model to fool multiple target networks. The fooling loss for training the generator is a linear combination of fooling losses of target models.

We consider two slightly different approaches for approximating $f(\cdot)$. The first approach is to parametrize it directly using a neural network $f_\Theta(\cdot)$. Hence, we seek $\Theta$ such that for most $x \in \mathfrak{N}$: $\mathcal{K}(f_\Theta(x)) \neq \mathcal{K}(x)$. We also require that the perturbed image $f_\Theta(x)$ look similar to the original image $x$. Hence, $d(x, f_\Theta(x))$ needs to be small for most $x \in \mathfrak{N}$, where $d(\cdot,\cdot)$ is a proper distance function. The second approach is to approximate the difference of natural and adversarial images with a neural network $f_\Theta(\cdot)$. We require that for most $x \in \mathfrak{N} : \mathcal{K}(x + f_\Theta(x)) \neq \mathcal{K}(x) \approx c_x$, and the $L_p$ norm of the additive perturbation $\|f_\Theta(x)\|_p$ needs to be small in order for it to be quasi-imperceptible. The second approach gives us better control over the perturbation magnitude. Hence, we will focus on this approach hereafter.

Figure 2 shows the architecture for generating image-dependent perturbations. Input image $x$ is passed through the generator to create the perturbation $f_\Theta(x)$. The perturbation is then scaled to constrain its norm. The result is the image-dependent perturbation which is added to the input image. We feed the clipped image $\hat{x}$ to the network to obtain the output probabilities $k(\hat{x})$. We use loss functions similar to the universal case as defined in equations 1–3. At inference time, we can discard the pre-trained model, and use only the generator to produce adversarial examples. This obviates the need for iterative gradient computations, and allows us to generate perturbations fast.

### 3.3. Fooling Multiple Networks

Using generative models for creating adversarial perturbations enables us to train sophisticated models. For instance, we can consider training a single model for misleading multiple networks simultaneously. Suppose we have models $\mathcal{K}_1, \mathcal{K}_2, \ldots, \mathcal{K}_m$ trained on natural images. Let $\mathcal{A}_\mathbf{K}$ denote the space of adversarial examples for these target models, i.e. $\mathcal{A}_\mathbf{K} = \{a \in [0,1]^n \backslash \mathfrak{N} \,|\, \exists\, x \in \mathfrak{N} : d(x,a) < \epsilon\,,\, \forall i \in \{1, \ldots, m\} : \mathcal{K}_i(a) \neq \mathcal{K}_i(x) \approx c_x\}$, in which $d(\cdot,\cdot)$ is a distance function, $\epsilon$ is a pre-specified threshold and $c_x$ is the ground-truth for $x$. We can consider both universal and image-dependent perturbations. In the case of universal perturbations, we seek a mapping $\mathcal{F} : [0,1]^n \rightarrow \mathcal{A}_\mathbf{K}$ generating adversarial examples from input patterns. In practice, the function is approximated with a deep neural network $\mathcal{F}_\Theta$. Figure 3 depicts the corresponding architecture. It is similar to figure 1 other than that the resulting perturbed image $\hat{x}$ is fed to each of the pre-trained models. The loss function for training the generator is a linear combination of fooling losses of pre-trained models as

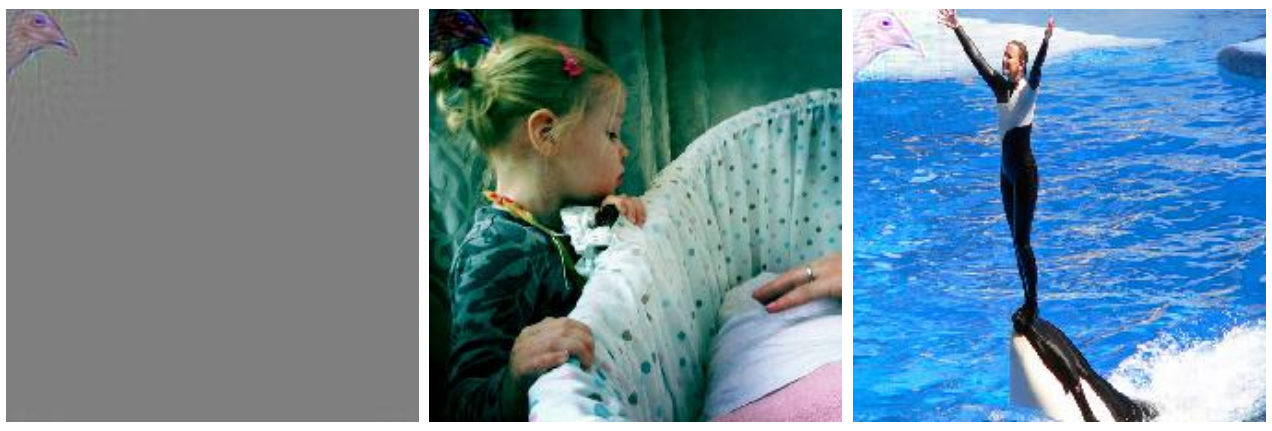

(a) Perturbation norm: $L_2 = 2000$, target model: VGG-16

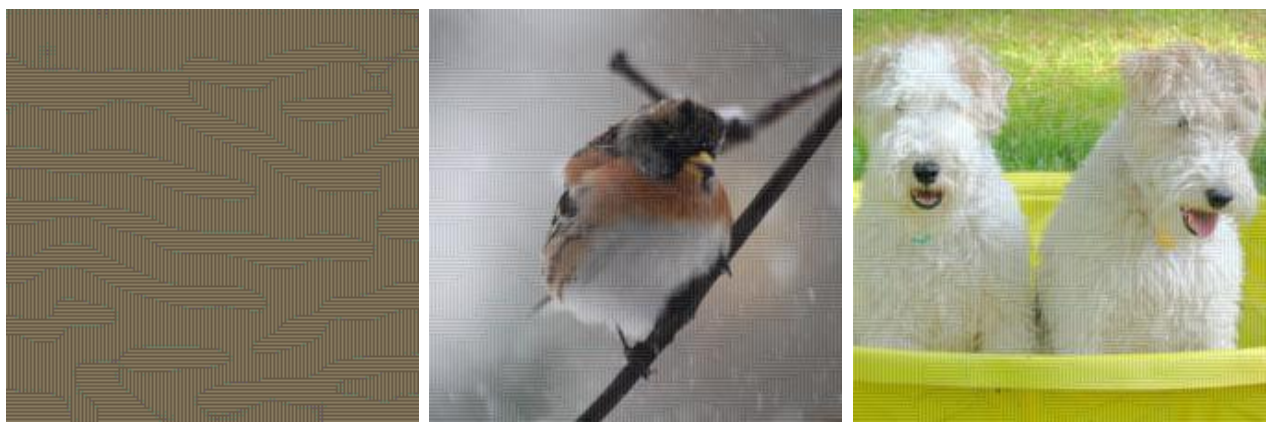

(b) Perturbation norm: $L_\infty = 10$, target model: VGG-19

Figure 4: Non-targeted universal perturbations. Enhanced universal pattern is shown on the left, and two samples of perturbed images are given on the right.

| | | VGG16 | VGG19 | ResNet152 |
|---|---|---|---|---|
| $L_2 = 2000$ | **GAP** | **93.9%** | **94.9%** | 79.5% |
| | UAP | 90.3% | 84.5% | **88.5%** |

Table 1: Fooling rates of non-targeted universal perturbations for various classifiers pre-trained on ImageNet. Our method (GAP) is compared with Universal Adversarial Perturbations (UAP) [35] using $L_2$ norm as the metric.

defined in equations 1–3. Hence, we have:

$$l_{multi-fool} = \lambda_1 \cdot l_{fool_1} + \cdots + \lambda_m \cdot l_{fool_m} \quad (4)$$

in which $\{\lambda_1, \ldots, \lambda_m\} \subset \mathbb{R}$ is a set of weights chosen based on the difficulty of deceiving each target model. The architecture for image-dependent perturbations is similar except that inputs to the generator are natural images.

# 4. Experiments on Classification

We generate adversarial examples for fooling classifiers pre-trained on the ImageNet dataset [14]. For the Euclidean distance as the metric, we scale the output of the generator to have a fixed $L_2$ norm. We can also scale the generator's output to constrain its maximum value when dealing with the $L_\infty$ norm. All results are reported on the 50,000 images of the ImageNet [14] validation set. Note that the contrast of displayed perturbations is enhanced for better visualization.

[7]Since [35] does not report results on Inception-v3, we compare with their results on Inception-v1 (GoogLeNet).

[7]This result uses the logit-based loss [7, 8] as opposed to the least-likely class loss (equation 2), which is used for other results in the table.

| | | VGG16 | VGG19 | Inception[6] |
|---|---|---|---|---|
| $L_\infty = 10$ | **GAP** | **83.7%** | **80.1%** | **82.7%**[7] |
| | UAP | 78.8% | 77.8% | 78.9% |

Table 2: Fooling rates of non-targeted universal perturbations using $L_\infty$ norm as the metric.

## 4.1. Universal Perturbations

**Non-targeted Universal Perturbations.** This setting corresponds to the architecture in figure 1 with the loss functions defined in equations 1 and 2. Results are given in Tables 1 and 2 for $L_2$ and $L_\infty$ norms respectively. For most cases our approach outperforms that of [35]. Similar to [35], a value of 2000 is set as the $L_2$-norm threshold of the universal perturbation, and a value of 10 is set for the $L_\infty$-norm when images are considered in [0, 255] range[8]. We use U-Net and ResNet Generator for $L_2$ and $L_\infty$ norms respectively. We visualize the results in figure 4. Notice that the $L_2$ perturbation consists of a bird-like pattern in the top left. Intuitively, the network has learned that in this constrained problem it can successfully fool the classifier for the largest number of images by converging to a bird perturbation. On the other hand, when we optimize the model based on $L_\infty$ norm, it distributes the perturbation to make use of the maximum permissible magnitude at each pixel.

**Targeted Universal Perturbations.** In this case we seek a single pattern which can be added to any image in the dataset to mislead the model into predicting a specified target label. We perform experiments with fixed $L_\infty$ norm of 10, and use the ResNet generator for fooling the Inception-v3 model. We use the loss function defined in equation 3 to train the generator. Figure 5 depicts the perturbations for various targets. It also shows the top-1 target accuracy on the validation set, i.e. the ratio of perturbed samples classified as the desired target. We observe the the universal perturbation contains patterns resembling the target class. While this task is more difficult than the non-targeted one, our model achieves high target accuracies. To the best of our knowledge, we are the first to present effective targeted universal perturbations on the ImageNet dataset. To make sure that the model performs well for any target, we train it on 10 randomly sampled classes. The resulting average target accuracy for $L_\infty = 10$ is 52.0%, demonstrating generalizability of the model across different targets.

## 4.2. Image-dependent Perturbations

[8] proposes a strong method for creating targeted image-dependent perturbations. However, its iterative al-

[8]The average $L_2$ and $L_\infty$ norm of images in our validation set are consistent with those reported in [35].

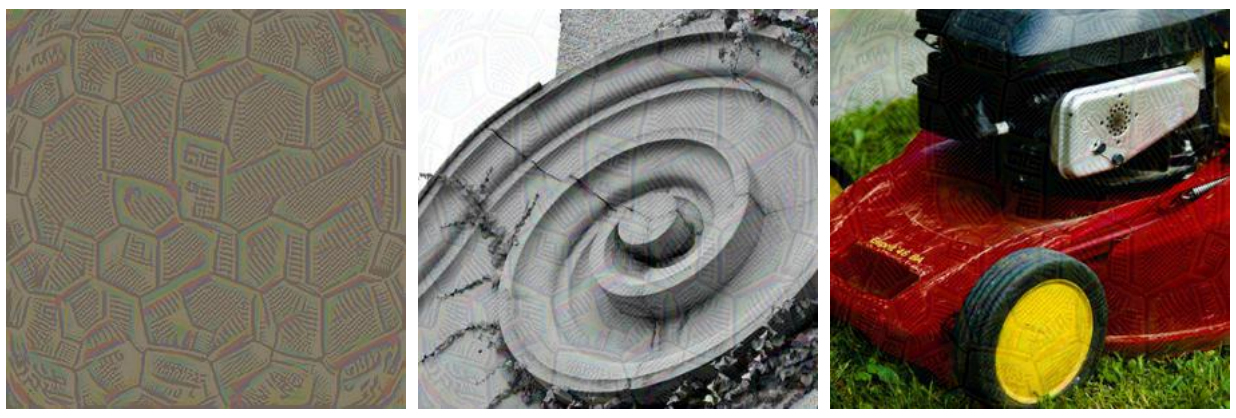

(a) Target: Soccer Ball, Top-1 target accuracy: 74.1%

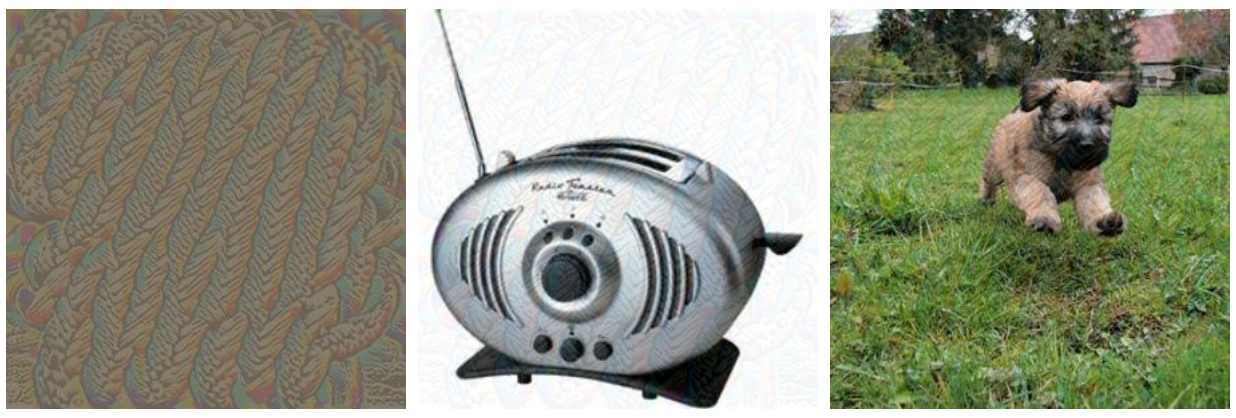

(b) Target: Knot, Top-1 target accuracy: 63.6%

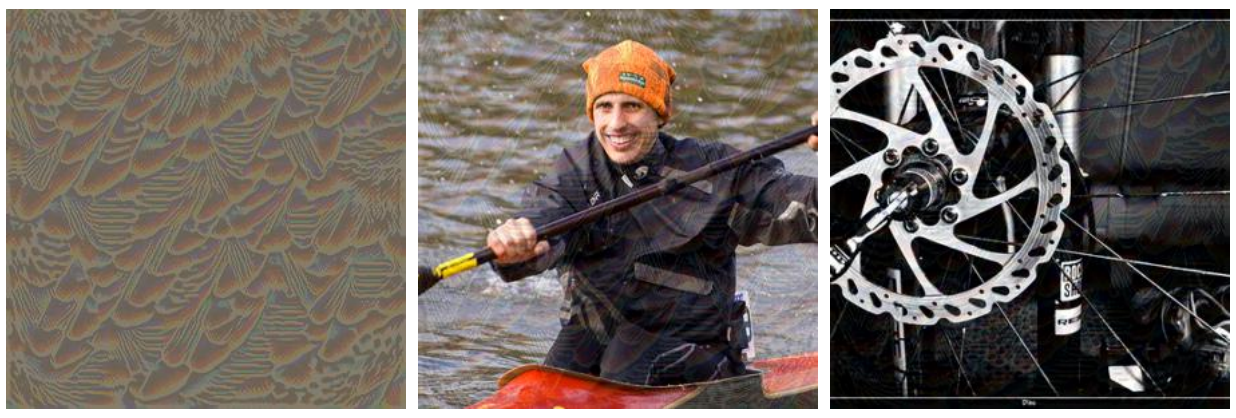

(c) Target: Finch, Top-1 target accuracy: 61.8%

Figure 5: Targeted universal perturbations. Three different targets and the corresponding average target accuracy of perturbed images on Inception-v3 are given. Universal pattern is shown on the left and two sample perturbed images are depicted on the right. Perturbation norm is $L_\infty = 10$.

gorithm is very slow at inference time. It reports attacks that take several minutes to run for each image, making it infeasible in real-time scenarios in which the input image changes constantly. FGSM [18] is a fast attack method but is not very accurate. In this work, we present adversarial attacks that are both fast and accurate.

**Non-targeted Image-dependent Perturbations.** The corresponding architecture is given in figure 2 with the loss function defined in equations 1 and 2. We use ResNet generator with 6 blocks for generating the perturbations. Similar to related works on image-dependent perturbations, we focus on $L_\infty$ norm as the metric. Results are shown for various perturbation norms and pre-trained classifiers in Table 3. Figure 6 illustrates the perturbed images. In this case the model converges to simple patterns which can change the prediction for most images. As we observe, the perturbations contain features from the corresponding input images.

**Targeted Image-dependent Perturbations.** For this task we use the training scheme shown in figure 2 with the loss function in equation 3. Figure 7 shows samples of perturbed images for fooling the Inception-v3 model. The perturbations are barely perceptible, yet they can obtain high target accuracies. Moreover, the perturbation itself has features resembling the target class and the input image. See figure 7 for more examples. We also evaluate performance of the model on 10 randomly sampled classes. The average target accuracy for $L_\infty = 10$ is $89.1\%$, indicating generalizability of the proposed model across different target classes. The average inference time for generating a perturbation to fool the Inception-v3 model is $0.28\ ms$ per image, showing that our method is considerably faster than [8][9].

| | $L_\infty = 7$ | $L_\infty = 10$ | $L_\infty = 13$ |
|---|---|---|---|
| VGG16 | 66.9% (30.0%) | 80.8% (17.7%) | 88.5% (10.6%) |
| VGG19 | 68.4% (28.8%) | 84.1% (14.6%) | 90.7% (8.6%) |
| Inception-v3 | 85.3% (13.7%) | 98.3% (1.7%) | 99.5% (0.5%) |

Table 3: Fooling ratios (pre-trained models' accuracies) for non-targeted image-dependent perturbations.

### 4.3. Transferability and Fooling Multiple Networks

Several works have demonstrated that adversarial examples generated for one model may also be misclassified by other models. This property is referred to as transferability, and can be leveraged to perform black-box attacks [53, 18, 40, 41, 29, 10, 7]. We show that our generated perturbations can be transferred across different models. Table 4 shows the fooling ratio of a non-targeted universal attack trained on one network and evaluated on others. Each row corresponds to the pre-trained model based on which the attack model is learned. The last row of the table corresponds to a model trained to jointly mislead VGG-16 and VGG-19 models based on the architecture depicted in figure 3. We see that joint optimization results in better transferability than training on a single target network. This is expected as the network has seen more models during training, so it generalizes better to unseen models.

## 5. Experiments on Semantic Segmentation

Current methods for fooling semantic segmentation models such as [58] and [34] use iterative algorithms, which are hand-engineered for the specific task, and are slow at inference. We demonstrate that our proposed architectures are generalizable across different tasks. More specifically, we show that architectures similar to those used in the classification task yield strong results on fooling segmentation models. We leave extension to tasks other than classification and segmentation as future work. Experiments are performed on the Cityscapes dataset [12]. It contains 2975 training and 500 validation images with a resolution of $2048 \times 1024$ pixels. Similar to [34], we downsample images and label maps to $1024 \times 512$ pixels using bilinear and nearest-neighbor interpolation respectively.

[9]The time is measured on Titan Xp GPUs.

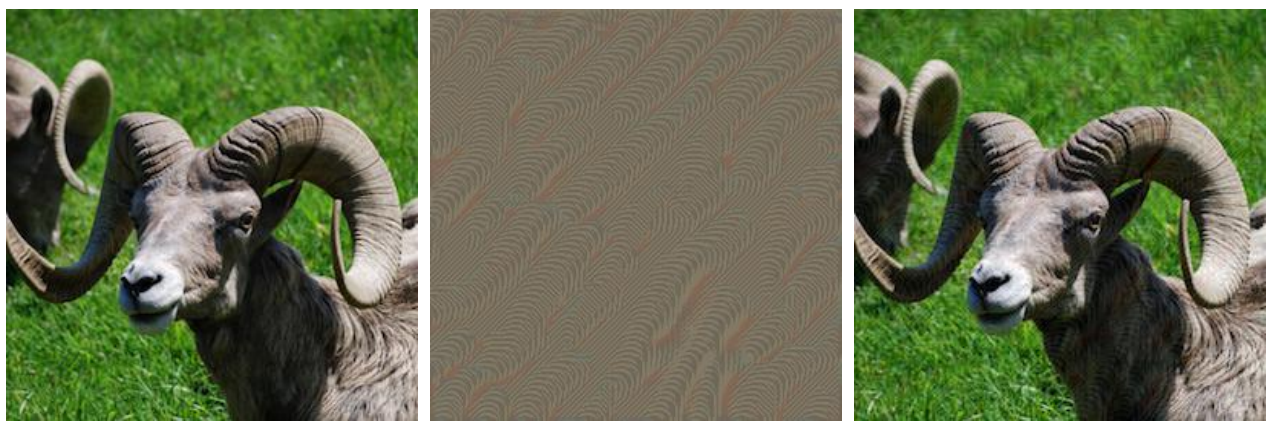

(a) $L_\infty = 7$

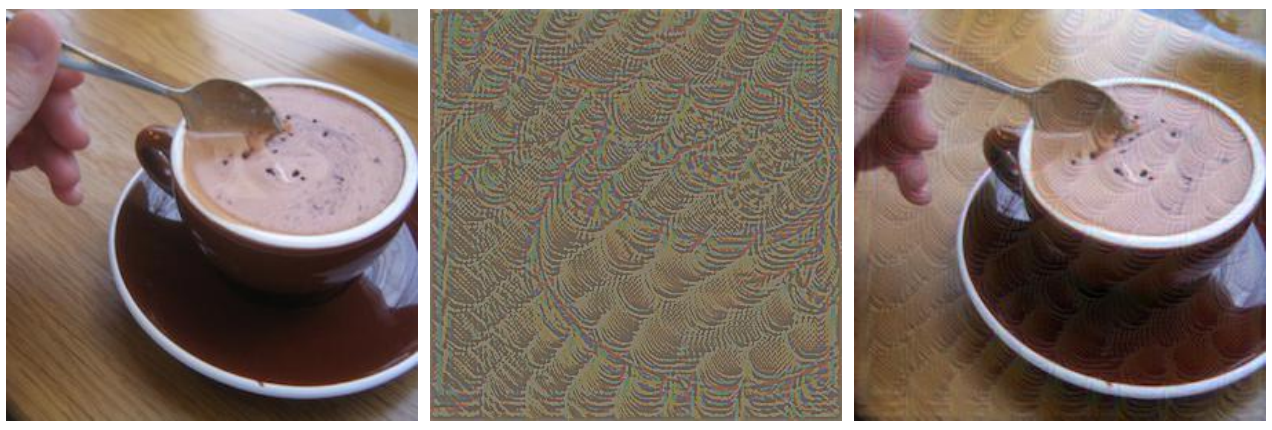

(b) $L_\infty = 10$

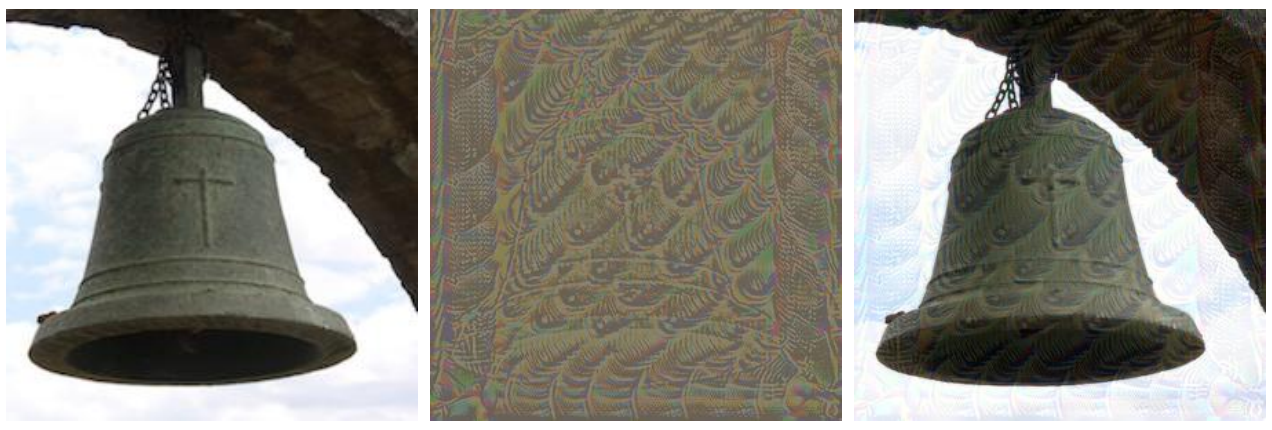

(c) $L_\infty = 13$

Figure 6: Non-targeted image-dependent perturbations. From left to right: original image, enhanced perturbation and perturbed image. Three different thresholds are considered with Inception-v3 as the target model.

| | VGG16 | VGG19 | ResNet152 |
|---|---|---|---|
| VGG16 | **93.9%** | 89.6% | 52.2% |
| VGG19 | 88.0% | **94.9%** | 49.0% |
| ResNet152 | 31.9% | 30.6% | **79.5%** |
| VGG16 + VGG19 | 90.5% | 90.1% | **54.1%** |

Table 4: Transferability of non-targeted universal perturbations. The network is trained to fool the pre-trained model shown in each row, and is tested on the model shown in each column. Perturbation magnitude is set to $L_2 = 2000$. The last row indicates joint training on VGG-16 and VGG-19.

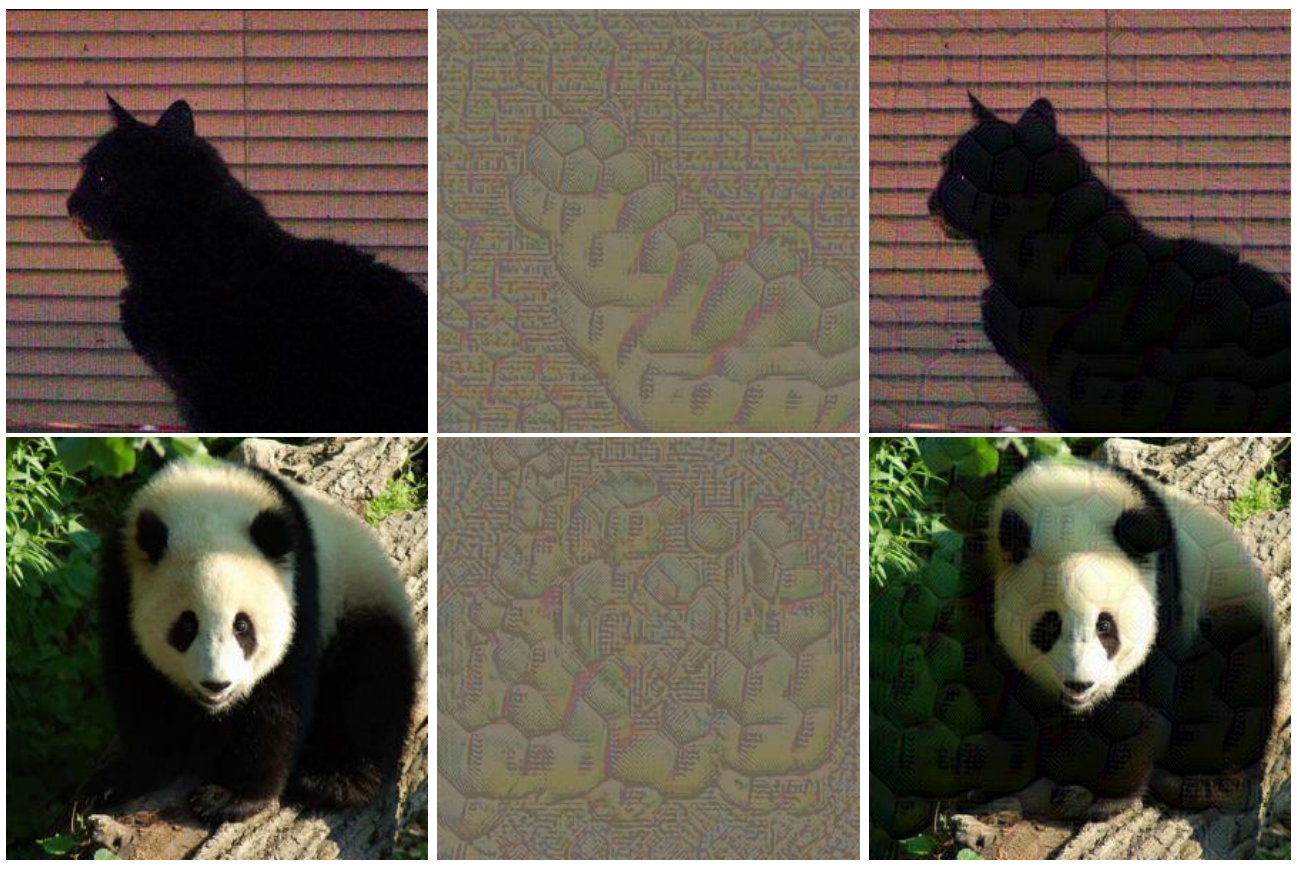

(a) Target: Soccer Ball, Top-1 target accuracy: 91.3%

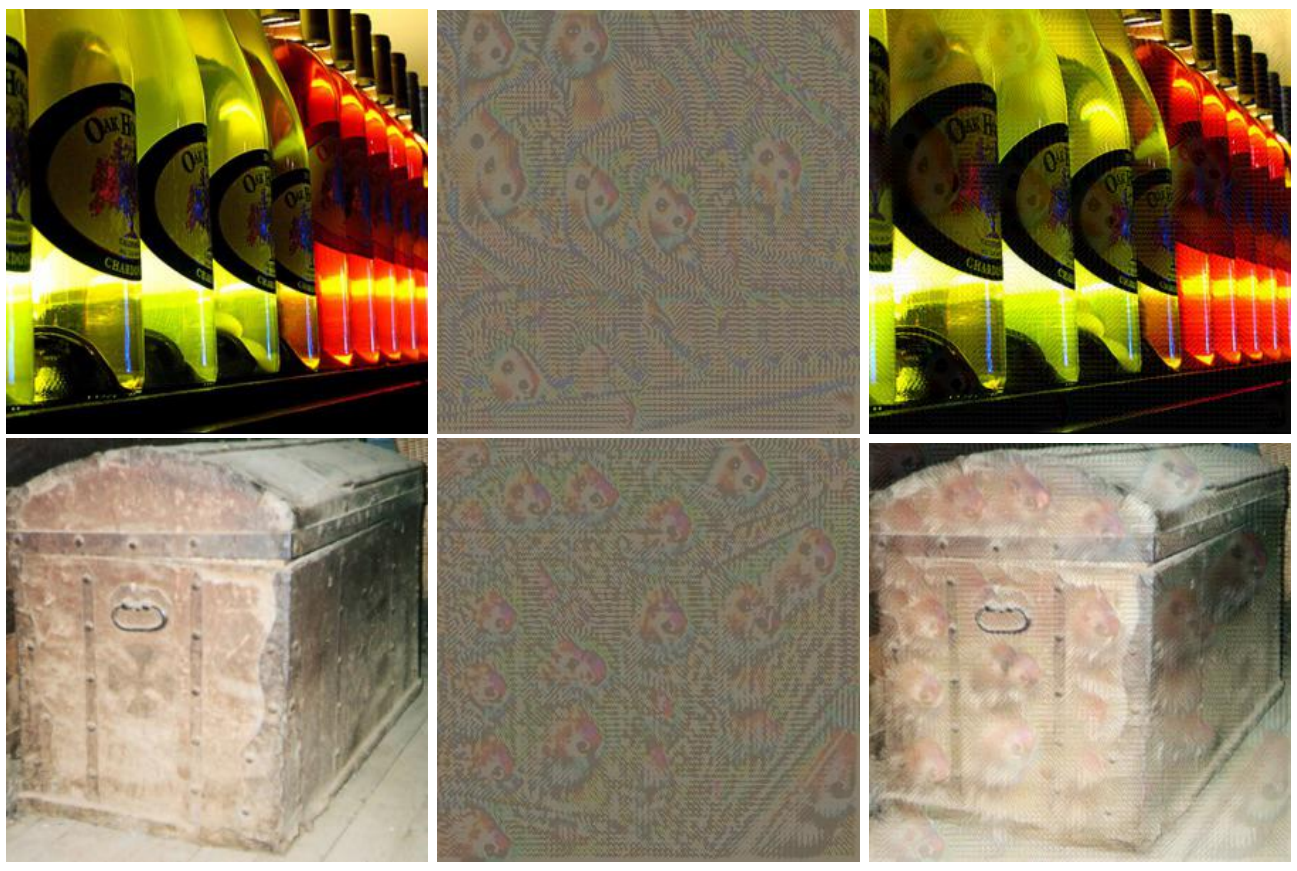

(b) Target: Hamster, Top-1 target accuracy: 87.4%

Figure 7: Targeted image-dependent perturbations. Two different targets and the corresponding average target accuracy of perturbed images on Inception-v3 are shown. From left to right: original image, enhanced perturbation and perturbed image. Perturbation magnitude is set to $L_\infty = 10$.

### 5.1. Universal Perturbations

We first consider the more challenging case of targeted attacks in which a desired target label map is given. We use the same setting as in the classification task, i.e. the training architecture in figure 1 with the fooling loss defined in equation 3. In order for our results to be comparable with [34], we consider FCN-8s [30] as our segmentation model, and use $L_\infty$ norm as the metric. Our setting corresponds to the *static target segmentation* in [34]. We use the same target as the paper, and consider our performance metric to be *success rate*, i.e. the categorical accuracy between the prediction $k(\hat{x})$ and the target $t$. Table 5 demonstrates our results. Our method outperforms the algorithm proposed in [34] for most of the perturbation norms. We also visualize the results in figure 8. We observe that the generator fools the segmentation model by creating a universal perturbation which resembles the target label map. We also demonstrates the resulting mean IoU for non-targeted attacks in Table 6.

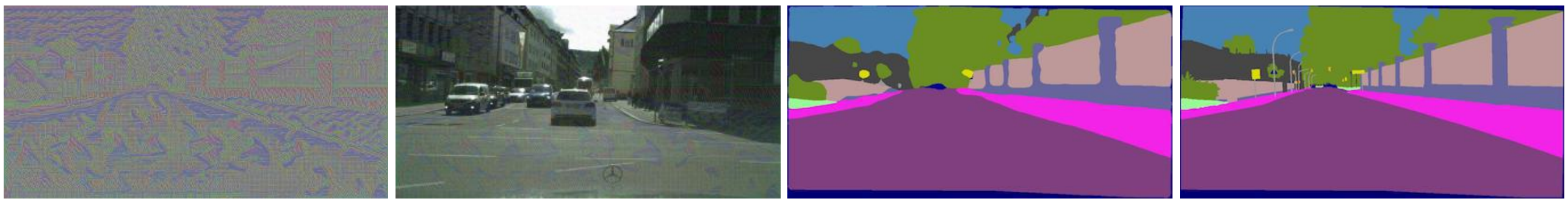

(a) Perturbation (b) Perturbed image (c) Prediction for perturbed image (d) Target

Figure 8: Targeted universal perturbations with $L_\infty = 10$ for fooling the FCN-8s semantic segmentation model.

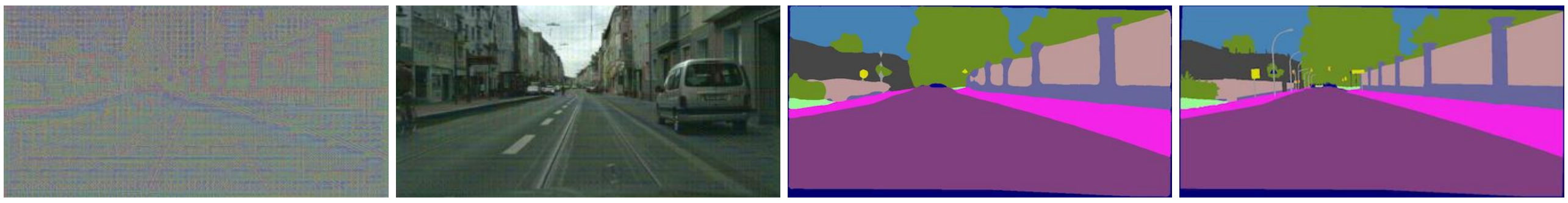

(a) Perturbation (b) Perturbed image (c) Prediction for perturbed image (d) Target

Figure 9: Targeted image-dependent perturbations with $L_\infty = 10$ for fooling the FCN-8s model.

| | $L_\infty = 5$ | $L_\infty = 10$ | $L_\infty = 20$ |
|---|---|---|---|
| **GAP (Ours)** | 79.5% | **92.1%** | **97.2%** |
| UAP-Seg [34] | **80.3%** | 91.0% | 96.3% |

Table 5: Success rate of targeted universal perturbations for fooling the FCN-8s segmentation model. Results are obtained on the validation set of the Cityscapes dataset.

| Task | $L_\infty = 5$ | $L_\infty = 10$ | $L_\infty = 20$ |
|---|---|---|---|
| Universal | 12.8% | 4.0% | 2.1% |
| Image-dependent | 6.9% | 2.1% | 0.4% |

Table 6: Mean IoU of non-targeted perturbations for fooling the FCN-8s segmentation model on the Cityscapes dataset.

### 5.2. Image-dependent Perturbations

The targeted image-dependent task corresponds to the architecture in figure 2 with the loss function in equation 3. We use the same target as the universal case. Results for various norms are given in Table 7. As we expect, relaxing the constraint of universality leads to higher success rates. Figure 9 illustrates the perturbations for $L_\infty = 10$. By closely inspecting the perturbations, we can observe patterns from both the target and the input image. As shown in Table 6, image-dependent perturbations achieve smaller mean IoU by not having the universality constraint. The average inference time per image is $132.82\ ms$ for the U-Net architecture and $335.73\ ms$ for the ResNet generator[10].

| | $L_\infty = 5$ | $L_\infty = 10$ | $L_\infty = 20$ |
|---|---|---|---|
| GAP | 87.0% | 96.3% | 98.2% |

Table 7: Success rate of targeted image-dependent perturbations for fooling FCN-8s on the Cityscapes dataset.

## 6. Discussion and Future Work

In this paper, we demonstrate the efficacy of generative models for creating adversarial examples. Four types of adversarial attacks are considered: targeted universal, non-targeted universal, targeted image-dependent and non-targeted image-dependent. We achieve high fooling rates on all tasks in the small perturbation norm regime. The perturbations can successfully transfer across different target models. Moreover, we demonstrate that similar architectures can be effectively used for fooling both classification and semantic segmentation models. This eliminates the need for designing task-specific attack methods, and paves the way for extending adversarial examples to other tasks. Future avenues of research include incorporating various properties such as transformation-invariance into the perturbations and extending the proposed framework to tasks other than classification and semantic segmentation.

[10] The time is measured on Titan Xp GPUs.

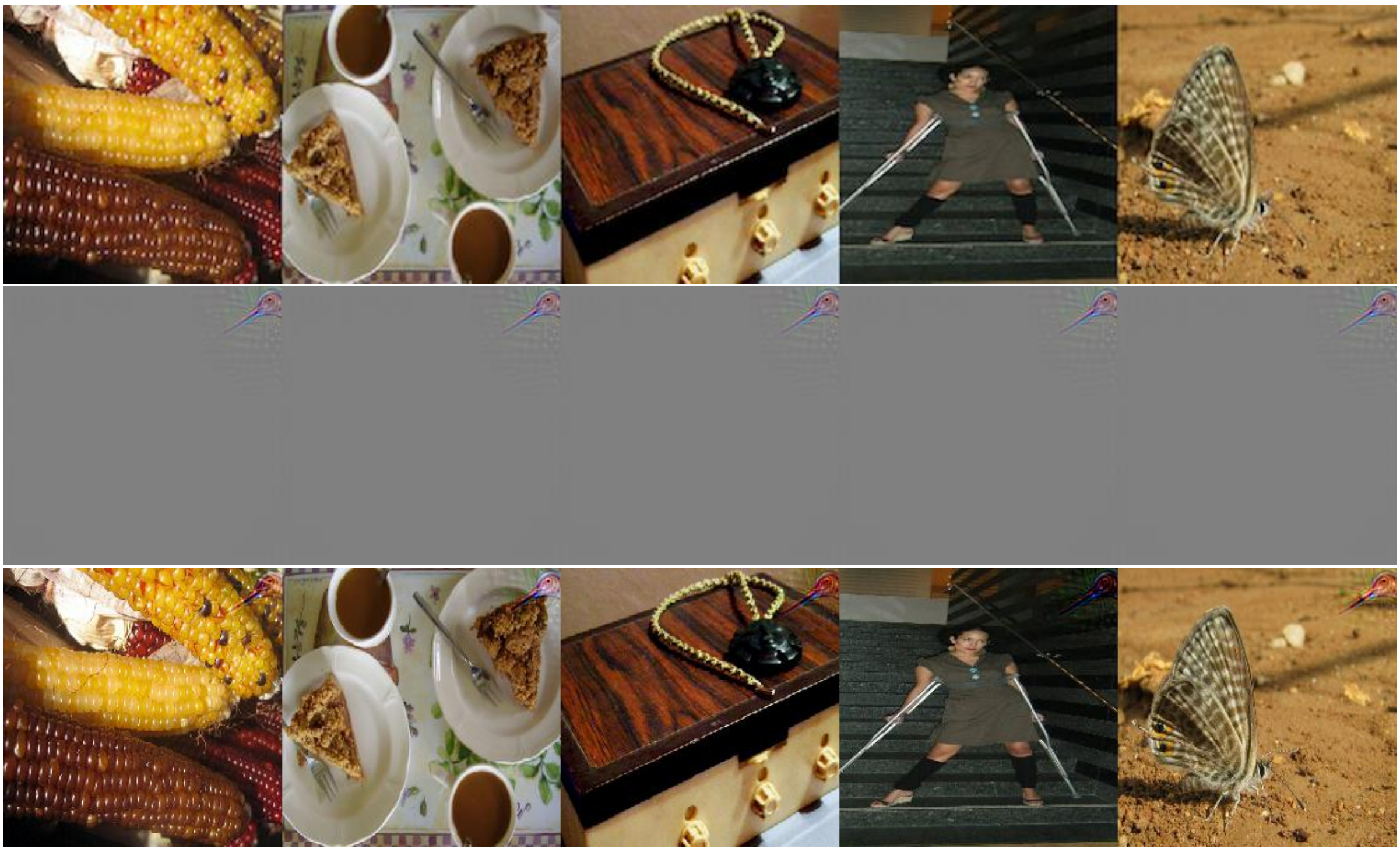

(a) Target model: VGG-19, Fooling ratio: 94.9%

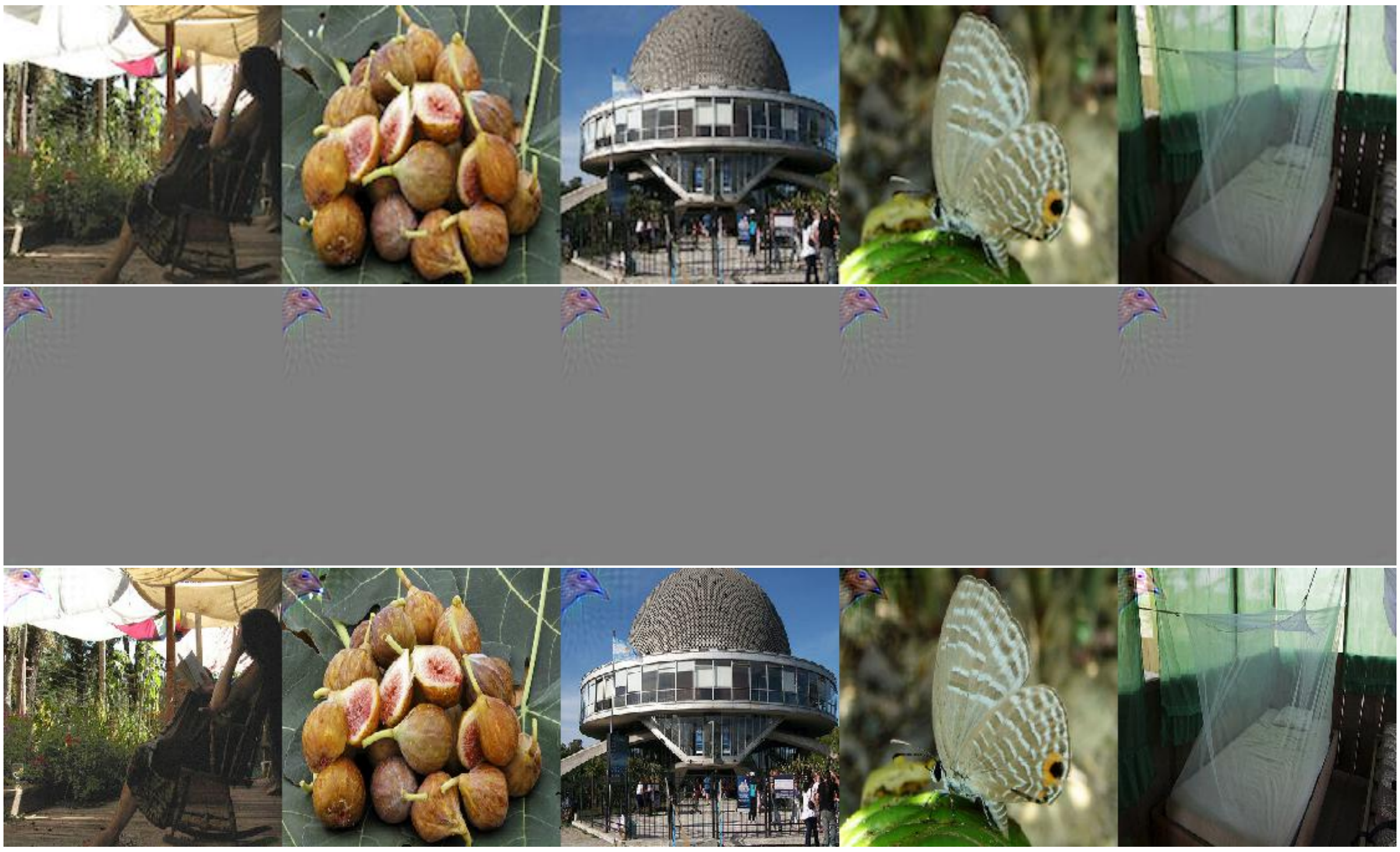

(b) Target model: VGG-16, Fooling ratio: 93.9%

Figure 10: Non-targeted universal perturbations. From top to bottom: original image, enhanced perturbation and perturbed image. Perturbation norm is set to $L_2 = 2000$ for (a) and (b) and to $L_\infty = 10$ for (c) and (d).

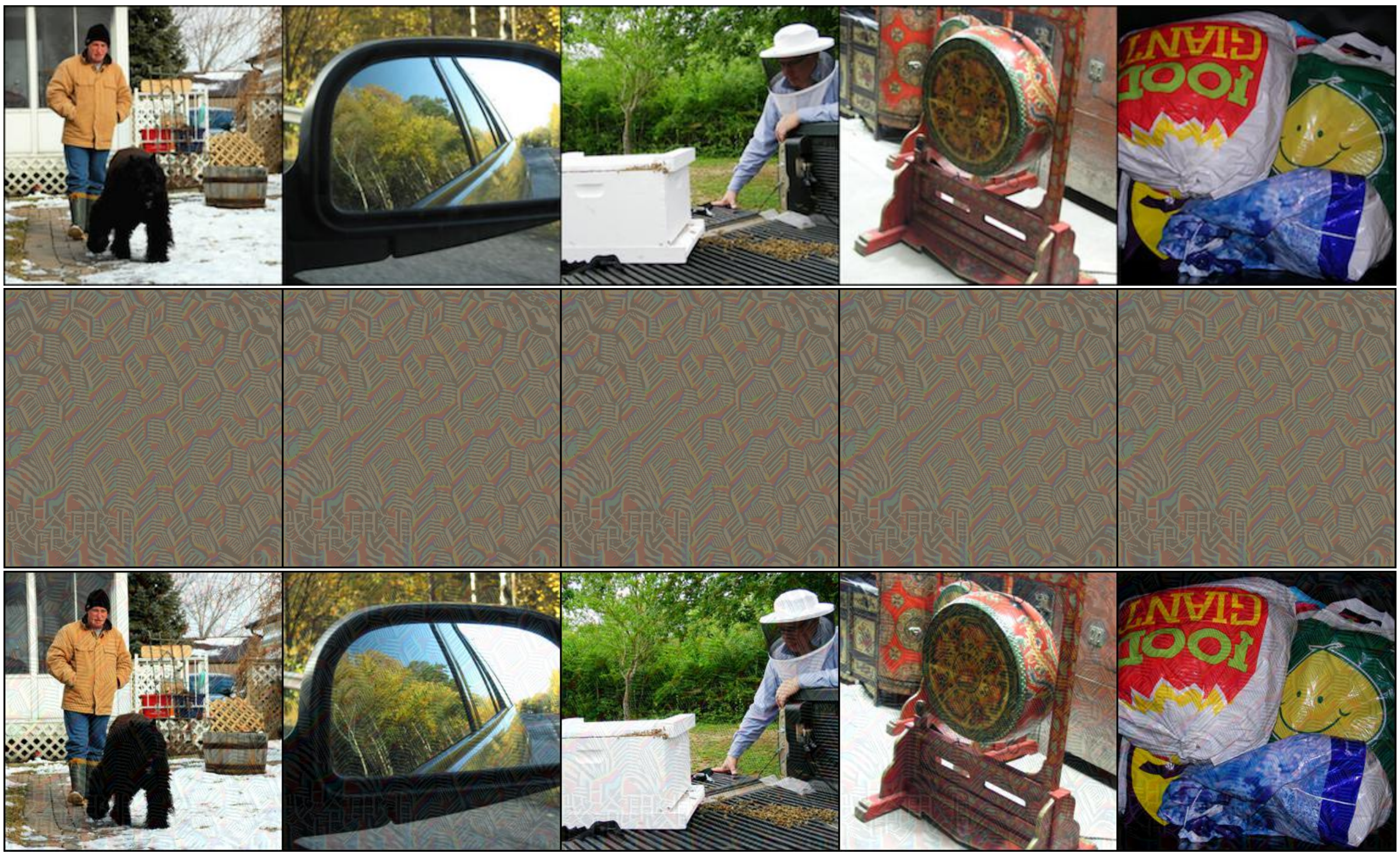

(c) Target model: Inception-v3, Fooling ratio: 79.2%

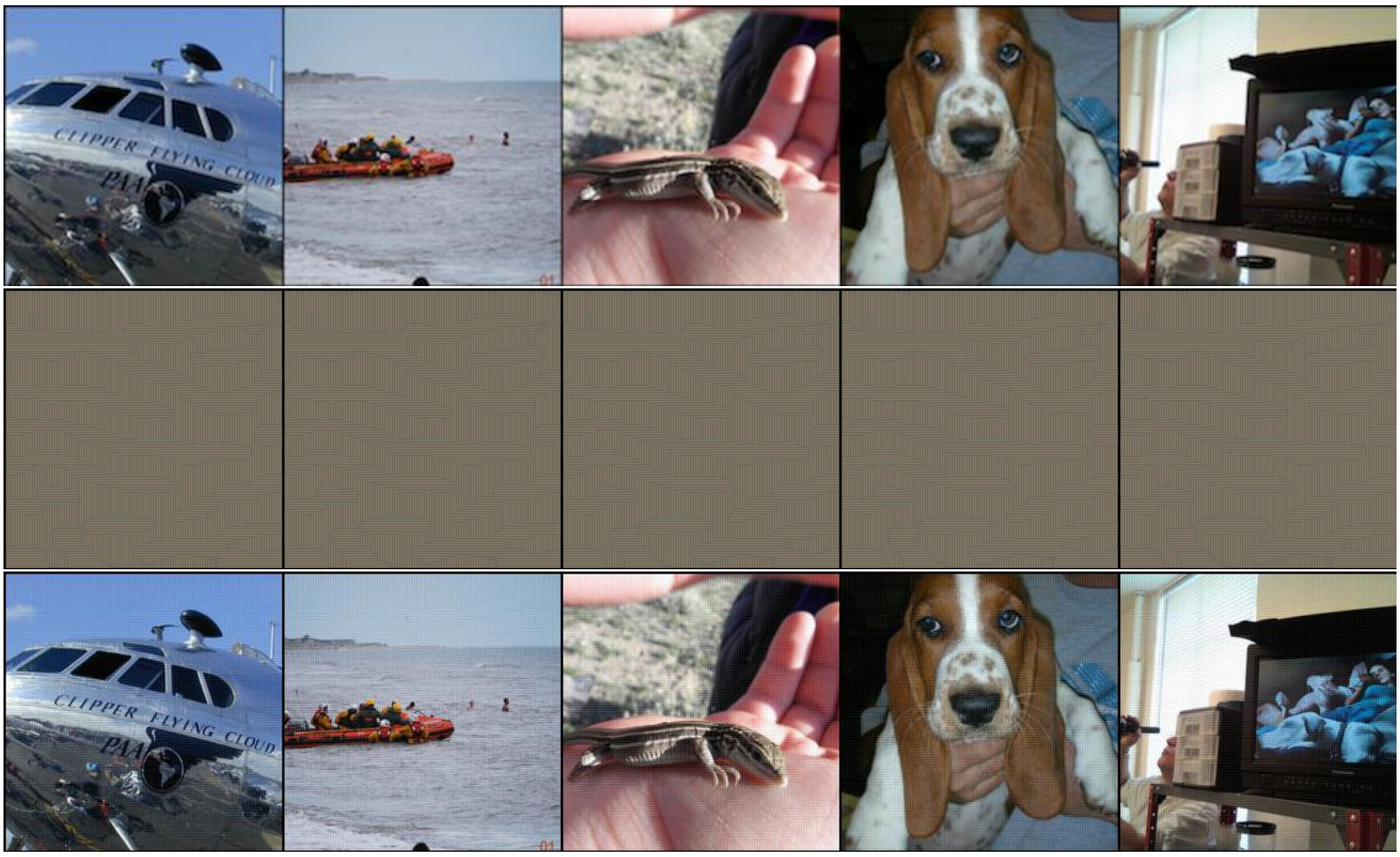

(d) Target model: VGG-19, Fooling ratio: 80.1%

Figure 10: Non-targeted universal perturbations (continued). From top to bottom: original image, enhanced perturbation and perturbed image. Perturbation norm is set to $L_2 = 2000$ for (a) and (b) and to $L_\infty = 10$ for (c) and (d).

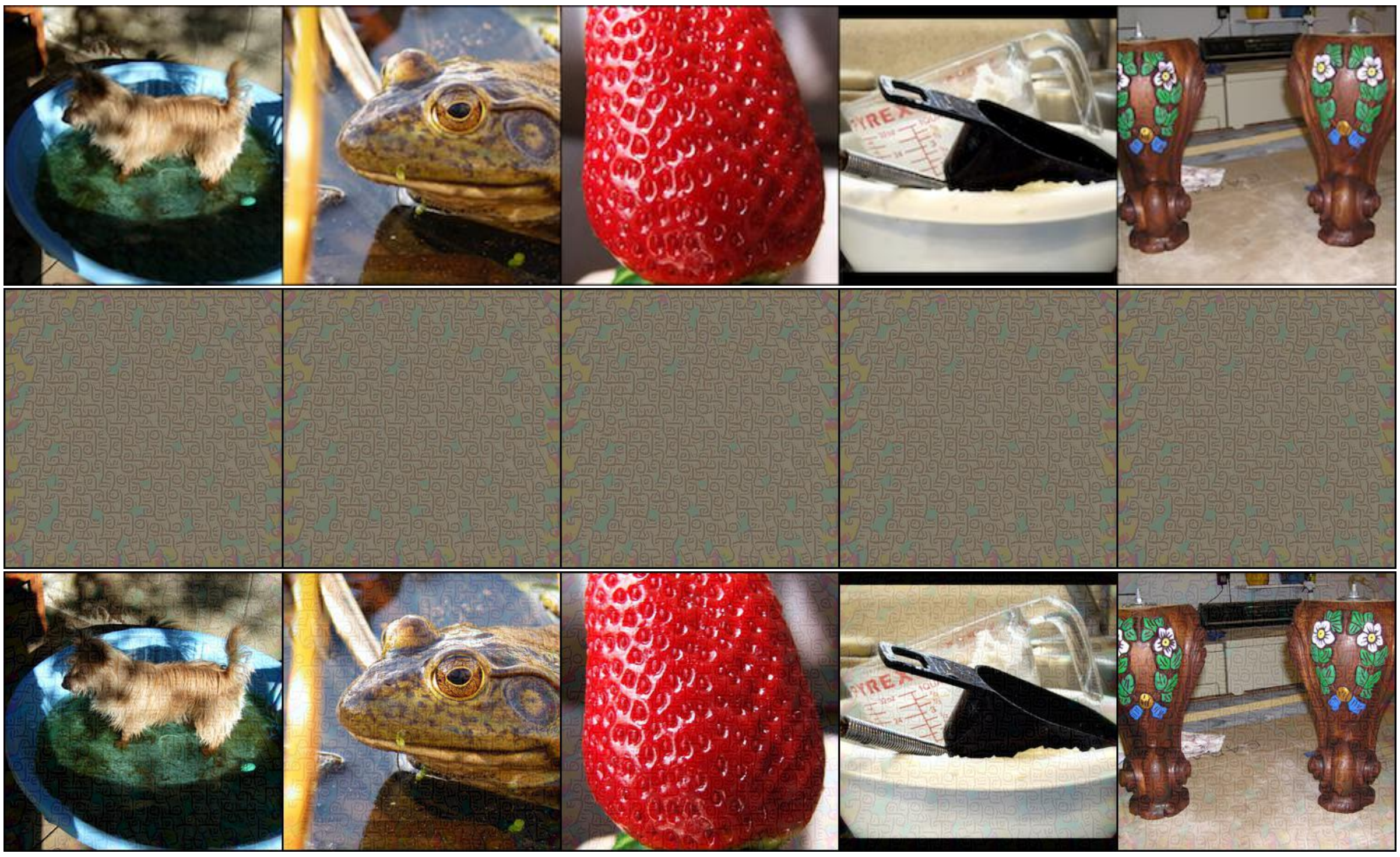

(a) Target: Jigsaw Puzzle, Top-1 target accuracy: 89.3%

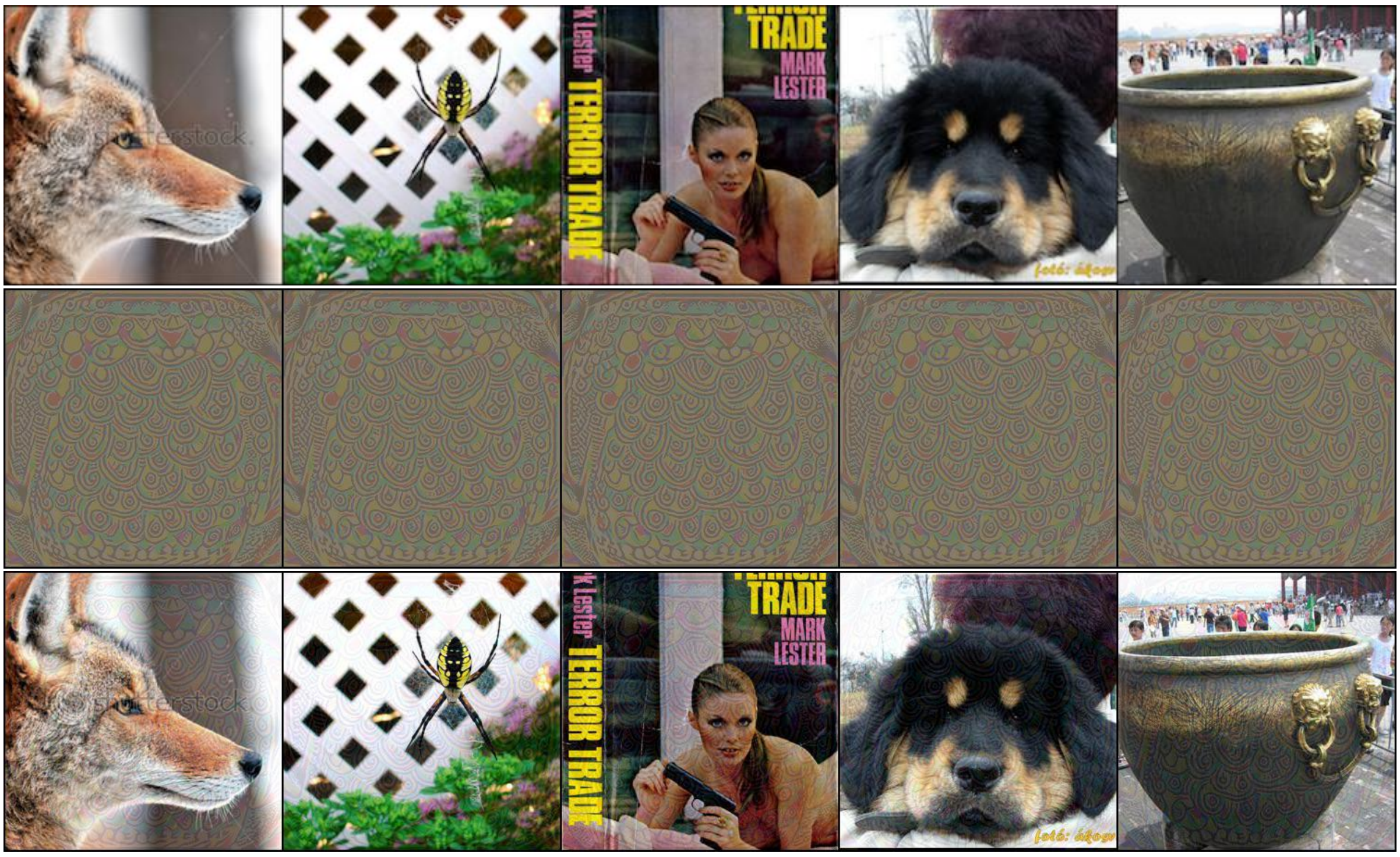

(b) Target: Teapot, Top-1 target accuracy: 62.2%

Figure 11: Targeted universal perturbations. From top to bottom: original image, enhanced perturbation and perturbed image. Perturbation norm is set to $L_\infty = 10$, and target model is Inception-v3.

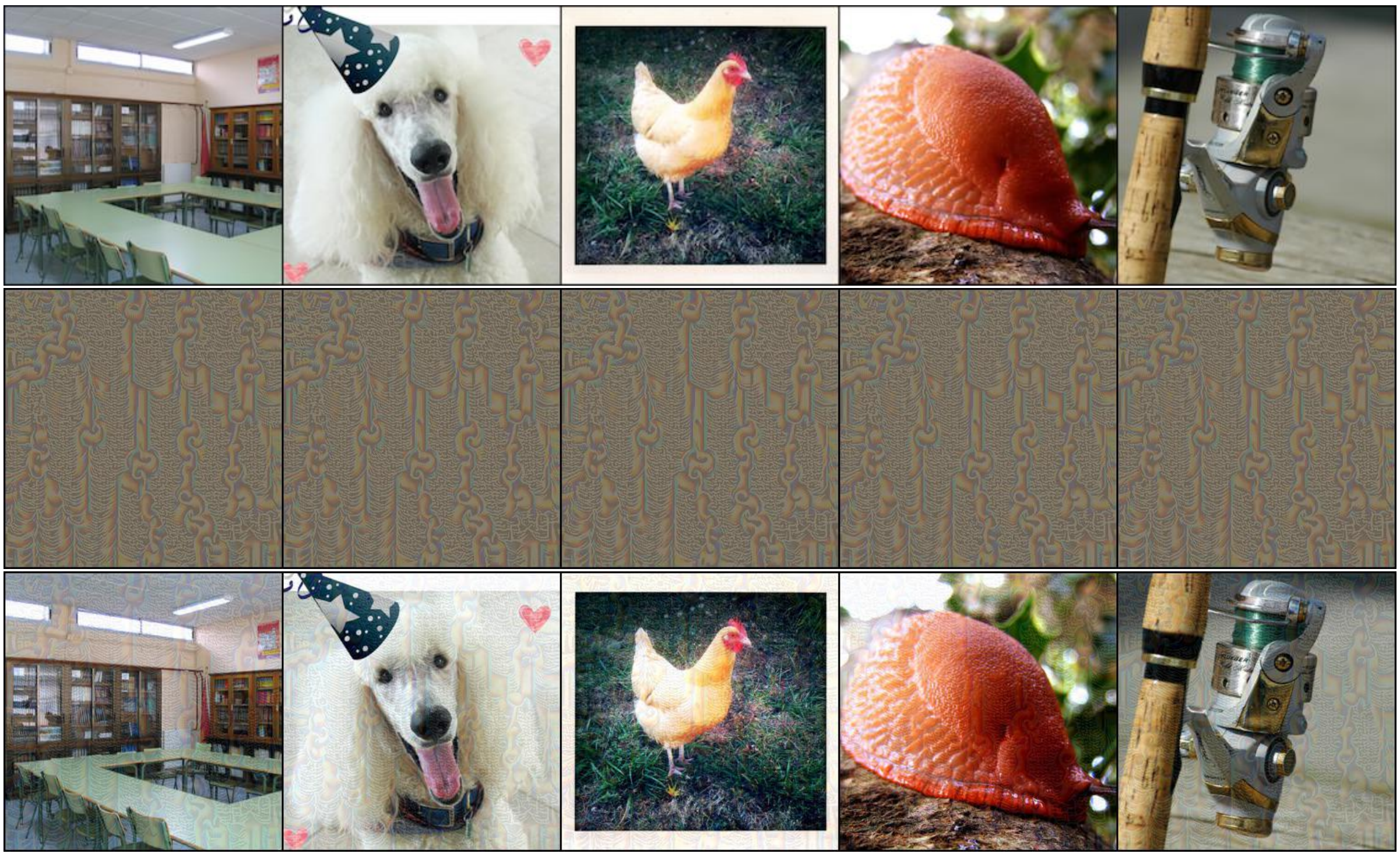

(c) Target: Chain, Top-1 target accuracy: 64.9%

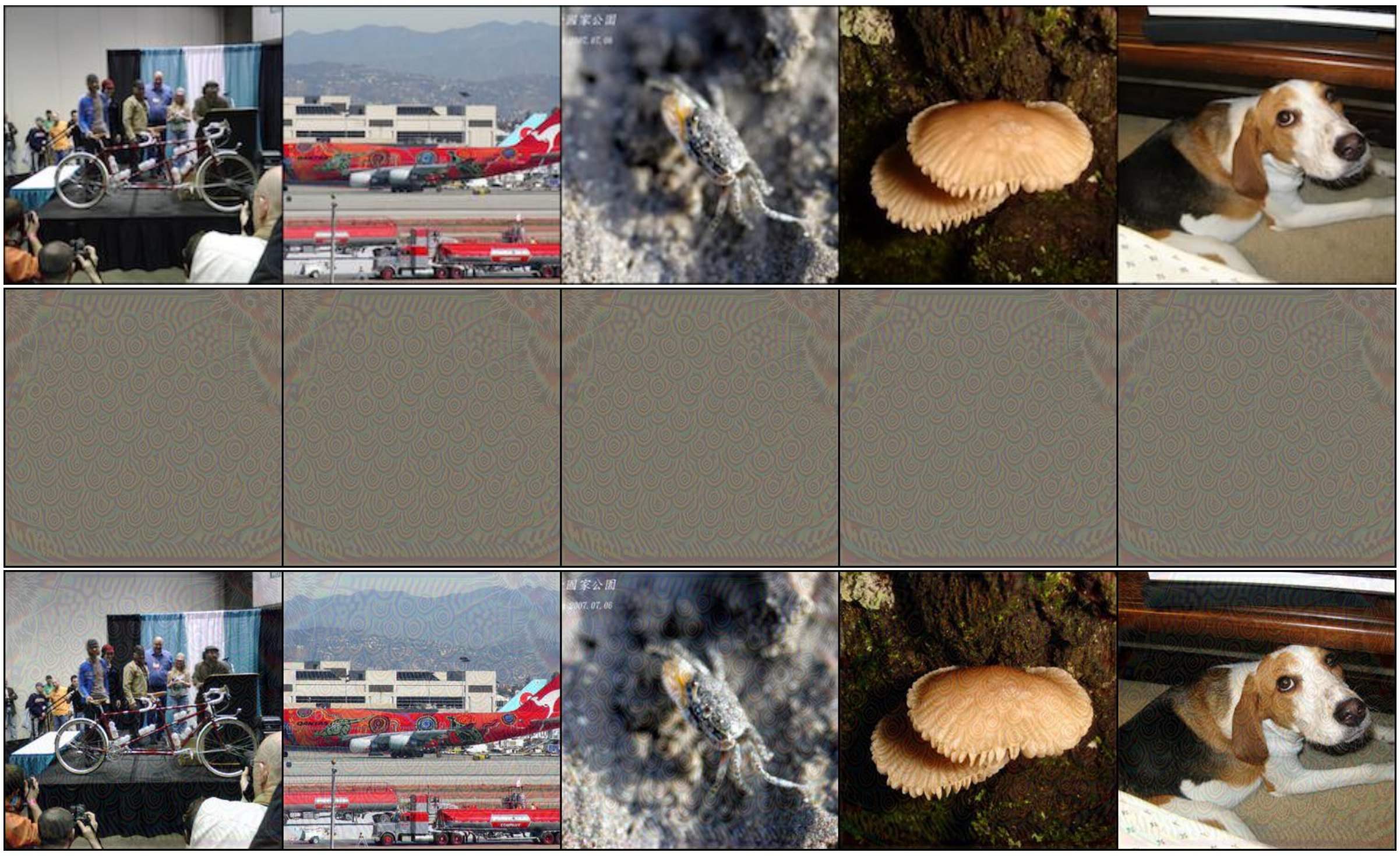

(d) Target: Hamster, Top-1 target accuracy: 60.0%

Figure 11: Targeted universal perturbations (continued). From top to bottom: original image, enhanced perturbation and perturbed image. Perturbation norm is set to $L_\infty = 10$, and target model is Inception-v3.

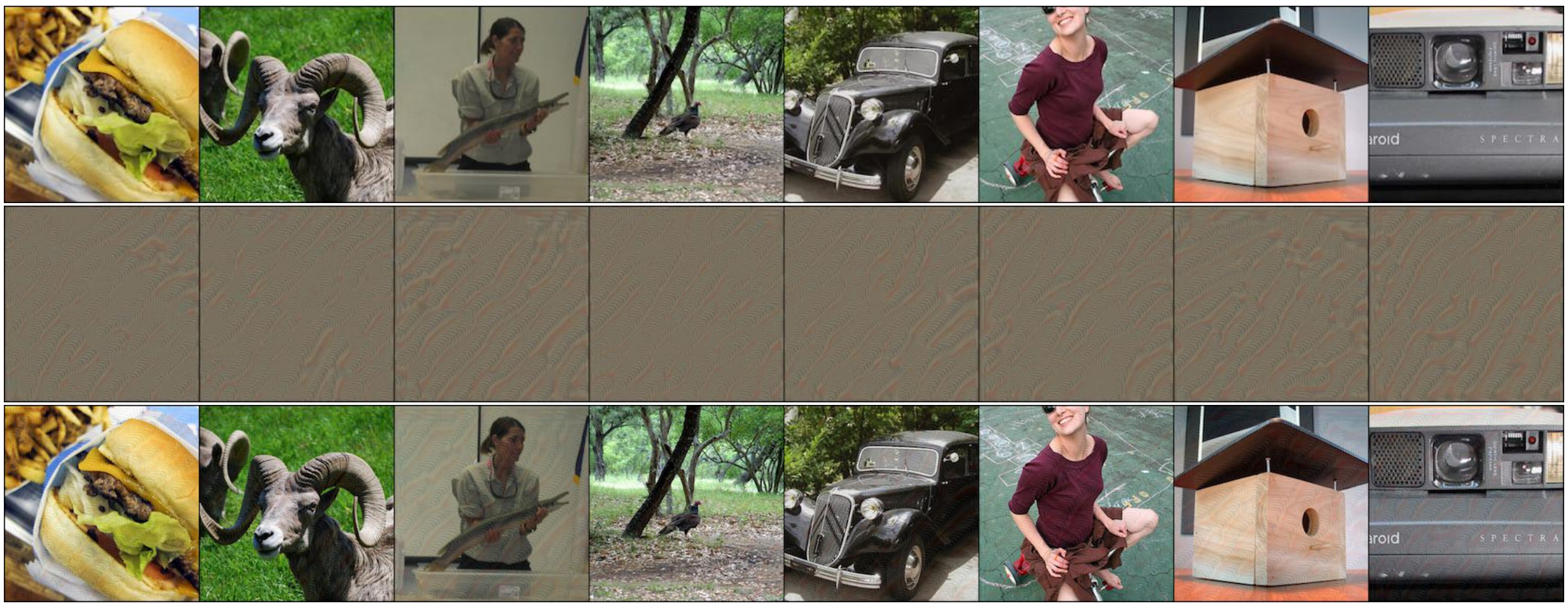

(a) $L_\infty = 7$

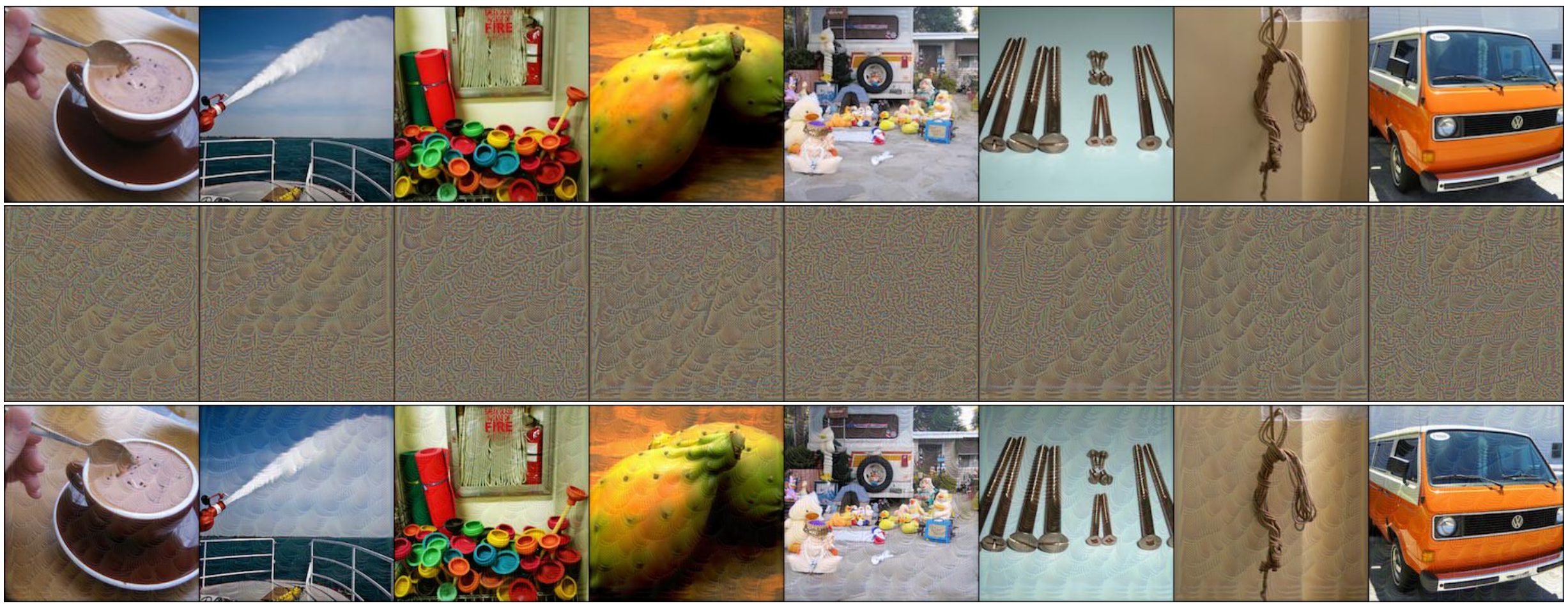

(b) $L_\infty = 10$

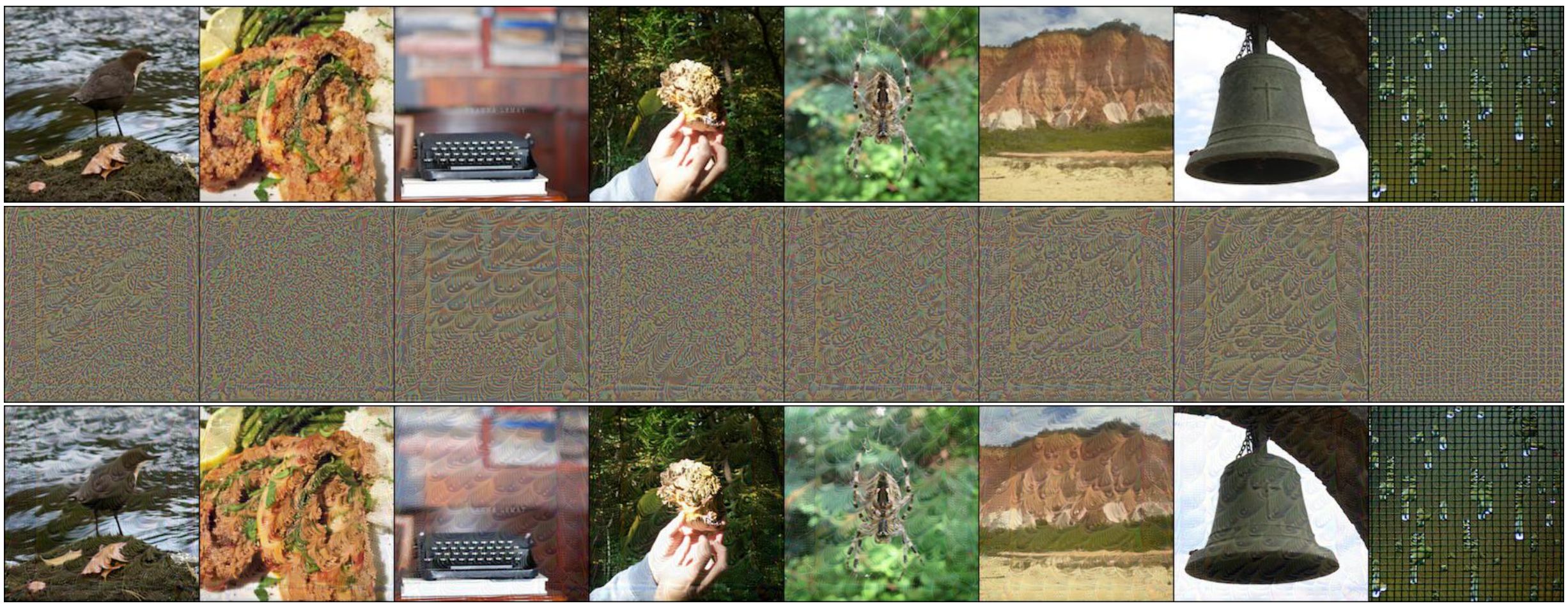

(c) $L_\infty = 13$

Figure 12: Non-targeted image-dependent perturbations. From top to bottom: original image, enhanced perturbation and perturbed image. Three different thresholds are considered with Inception-v3 as the target model.

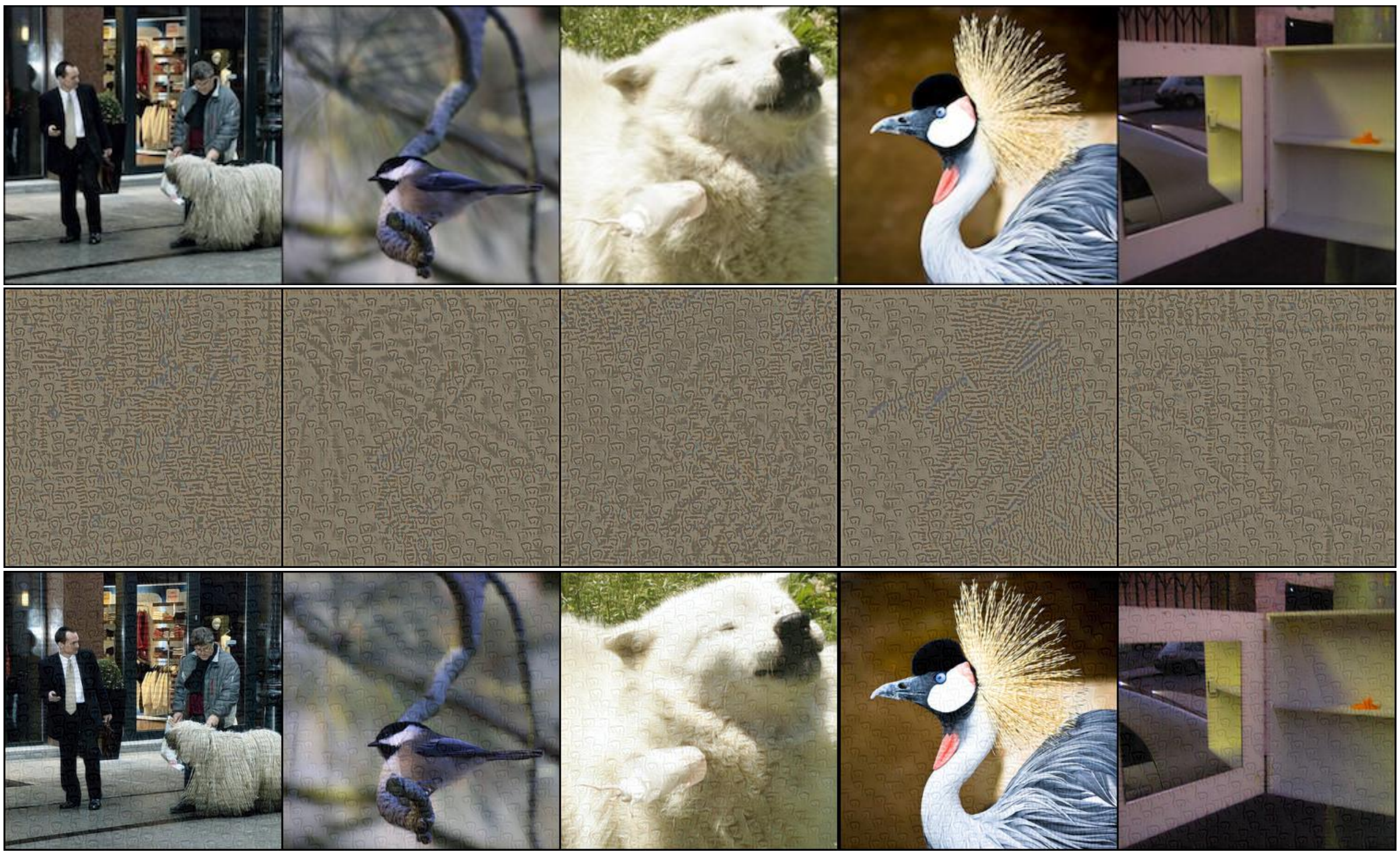

(a) Target: Jigsaw puzzle, Top-1 target accuracy: 98.1%

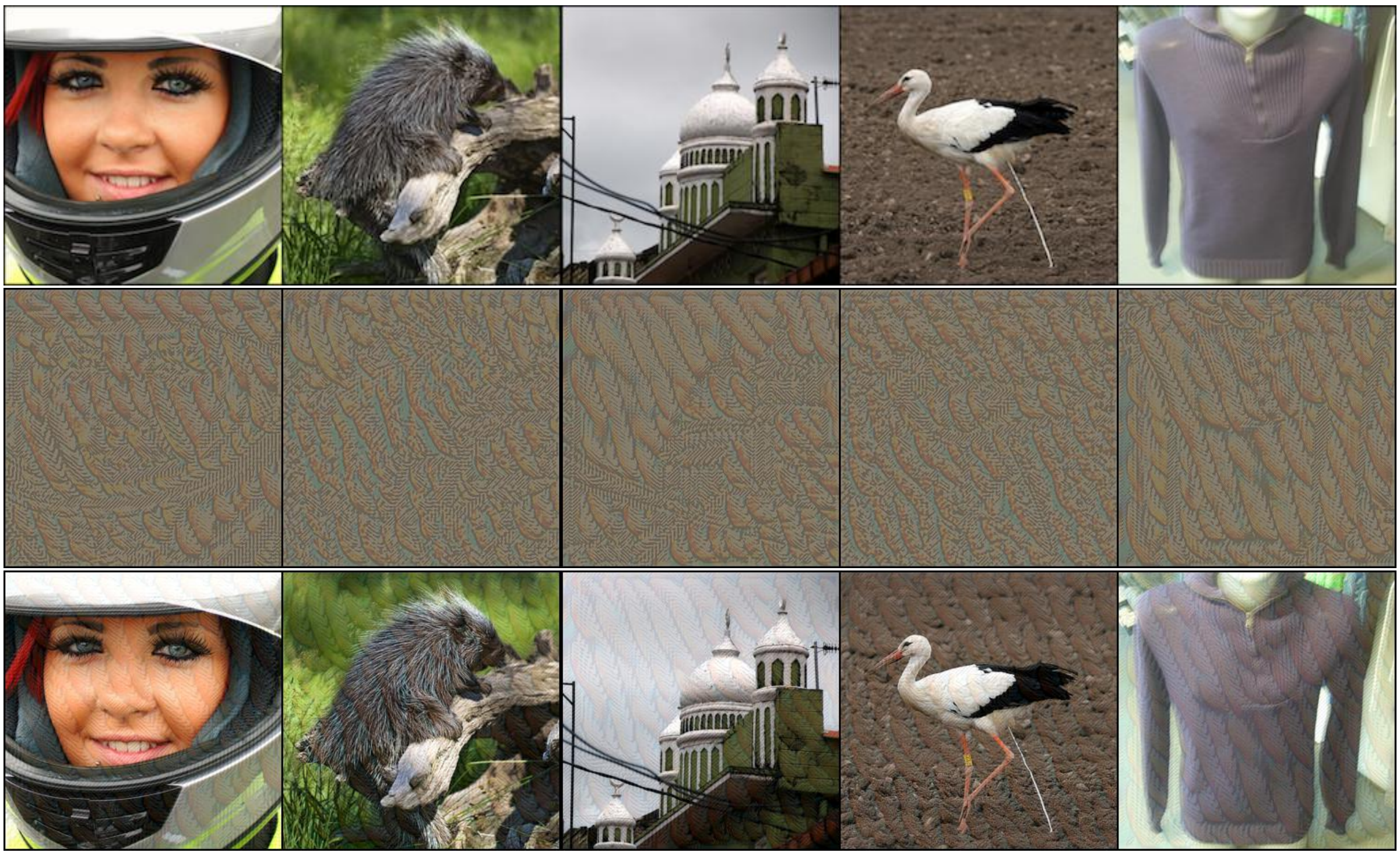

(b) Target: Knot, Top-1 target accuracy: 95.0%

Figure 13: Targeted image-dependent perturbations. From top to bottom: original image, enhanced perturbation and perturbed image. Perturbation norm is set to $L_\infty = 10$, and Inception-v3 is the pre-trained model.

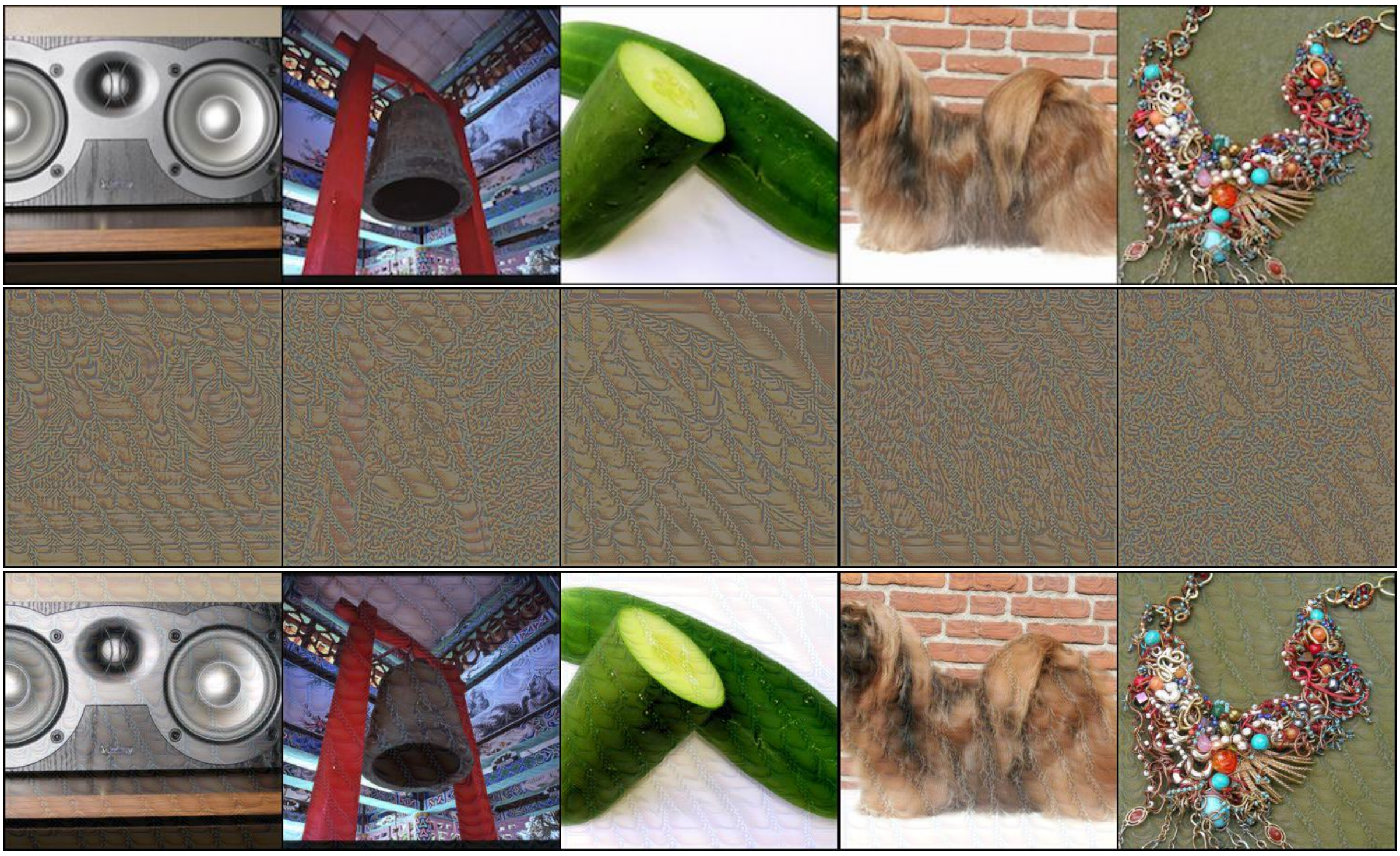

(c) Target: Chain, Top-1 target accuracy: 89.7%

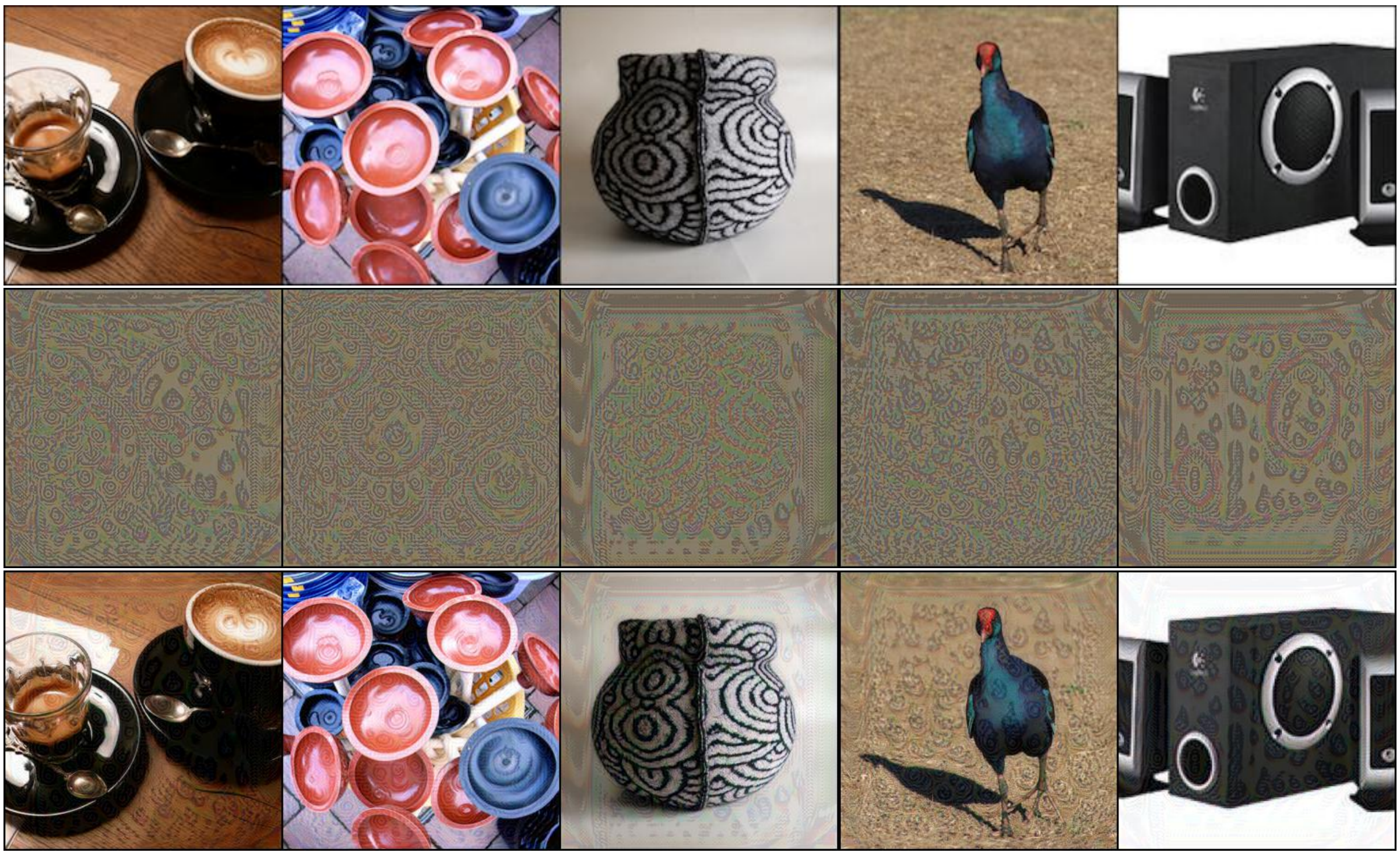

(d) Target: Teapot, Top-1 target accuracy: 90.6%

Figure 13: Targeted image-dependent perturbations (continued). From top to bottom: original image, enhanced perturbation and perturbed image. Perturbation norm is set to $L_\infty = 10$, and Inception-v3 is the pre-trained model.

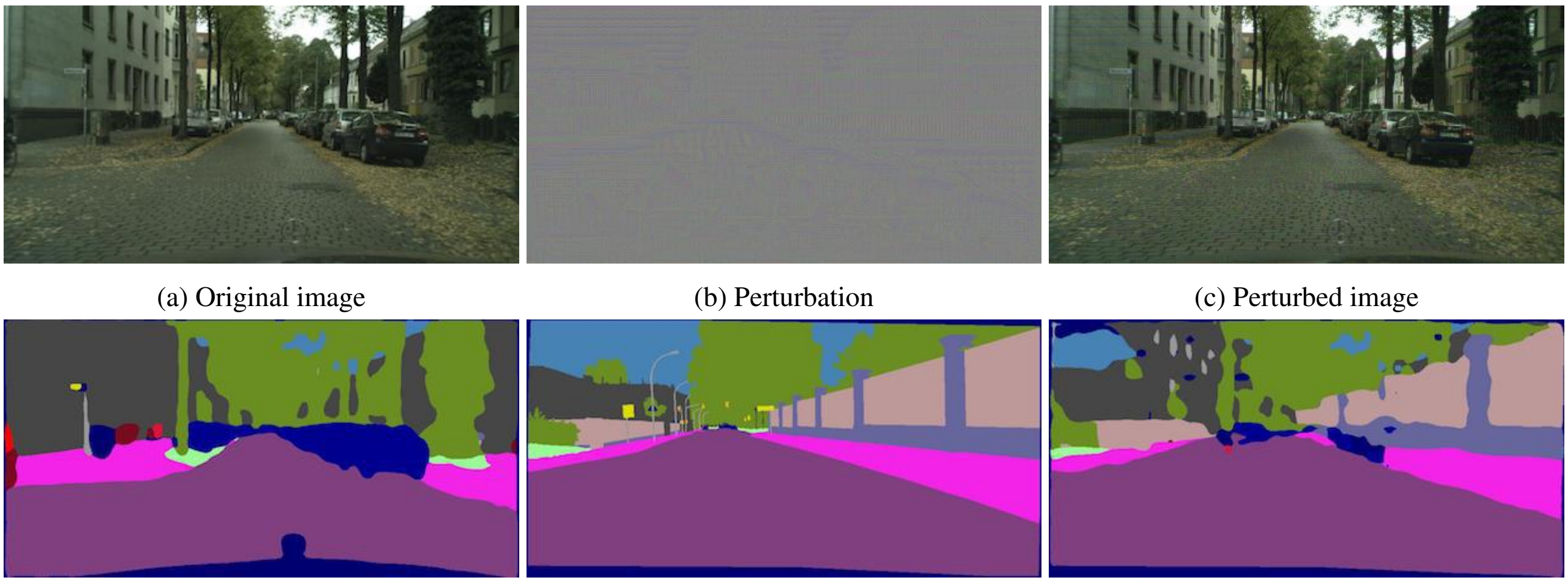

(a) Original image (b) Perturbation (c) Perturbed image

(d) Prediction for original image (e) Target (f) Prediction for perturbed image

Figure 14: Targeted universal perturbations with $L_\infty = 5$. Zoom in for details.

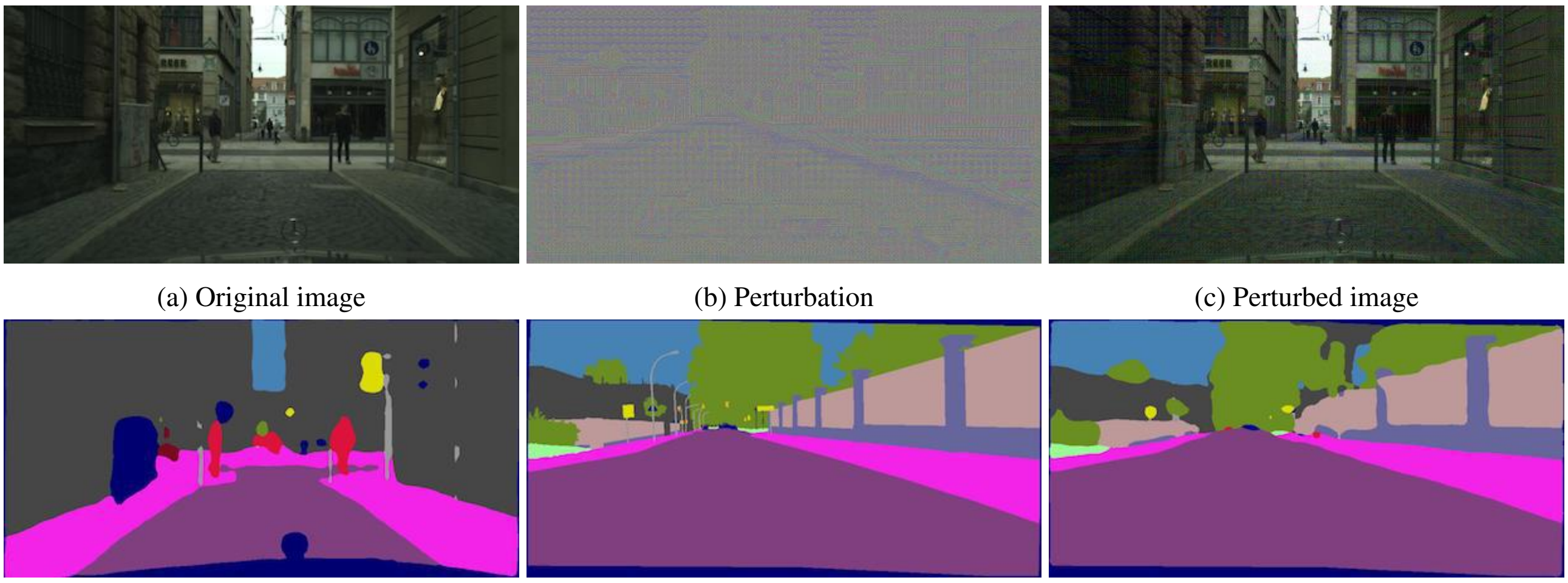

(a) Original image (b) Perturbation (c) Perturbed image

(d) Prediction for original image (e) Target (f) Prediction for perturbed image

Figure 15: Targeted universal perturbations with $L_\infty = 10$.

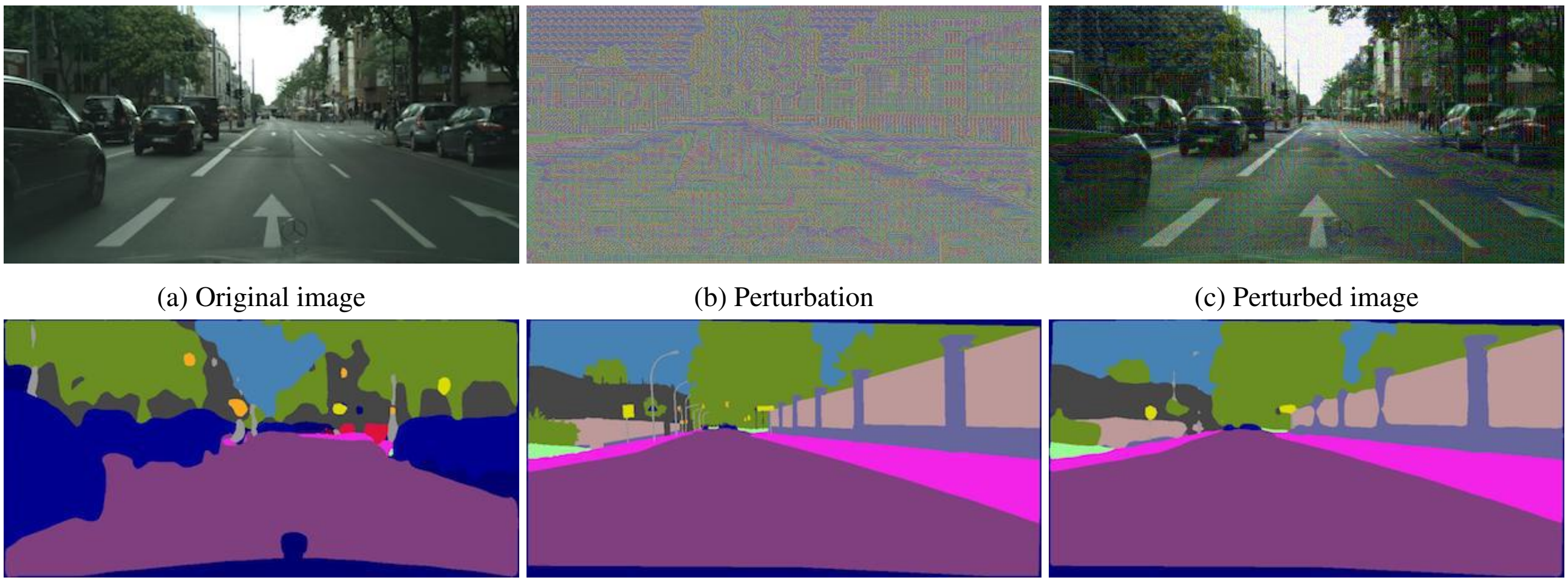

(a) Original image (b) Perturbation (c) Perturbed image

(d) Prediction for original image (e) Target (f) Prediction for perturbed image

Figure 16: Targeted universal perturbations with $L_\infty = 20$.

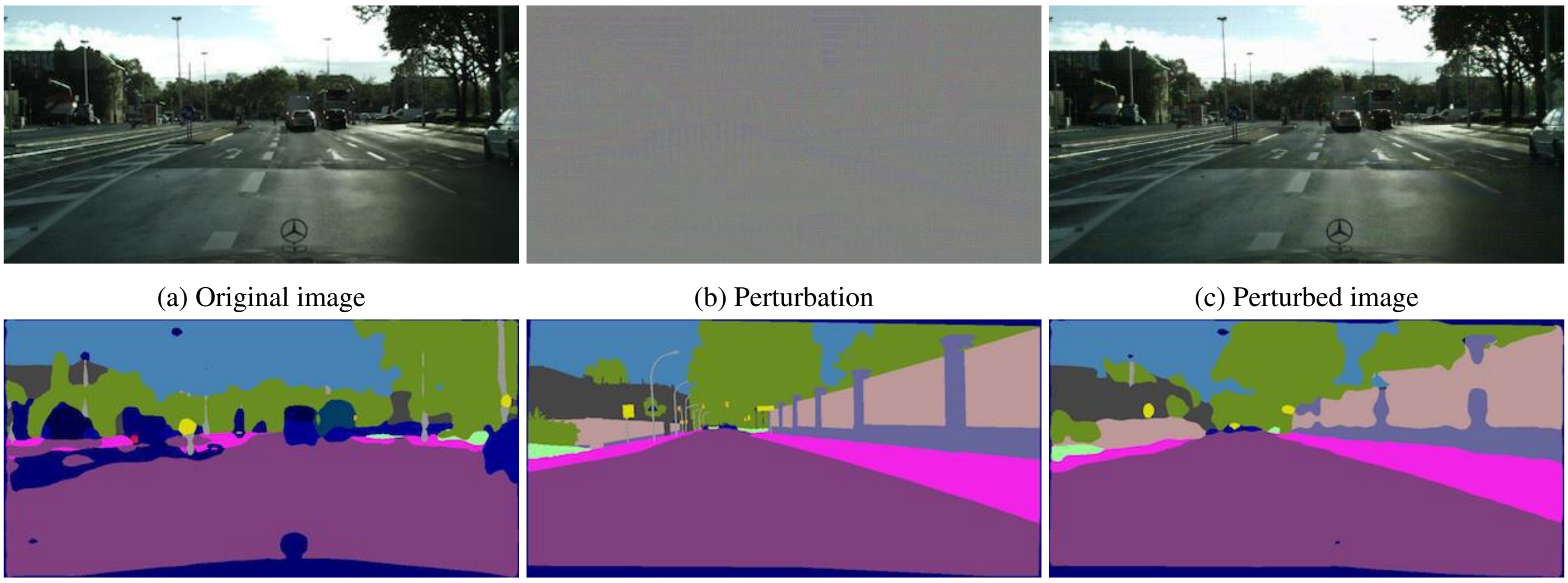

(a) Original image (b) Perturbation (c) Perturbed image

(d) Prediction for original image (e) Target (f) Prediction for perturbed image

Figure 17: Targeted image-dependent perturbations with $L_\infty = 5$. Zoom in for details.

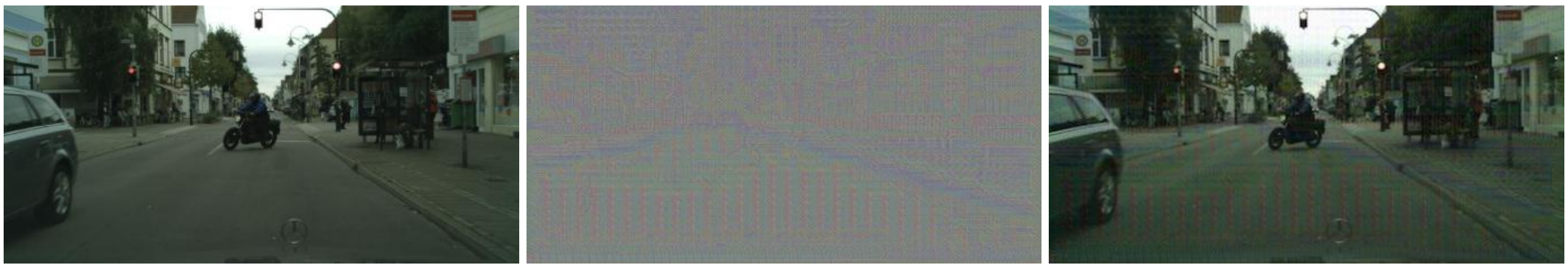

(a) Original image (b) Perturbation (c) Perturbed image

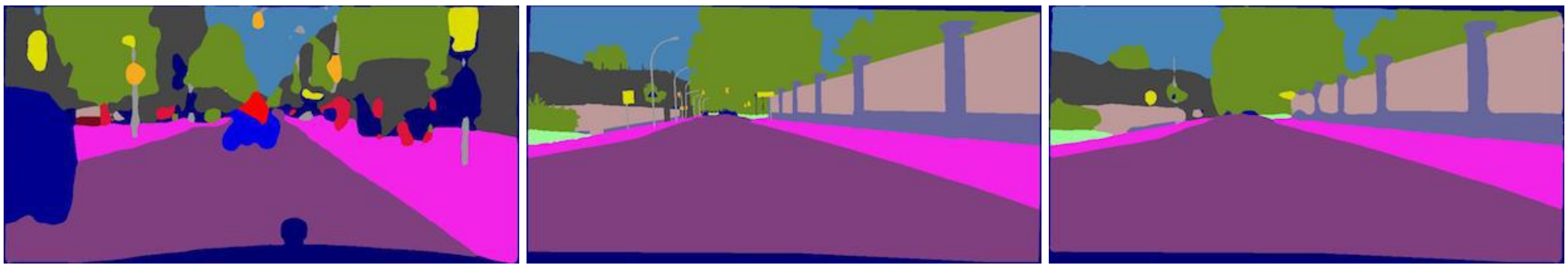

(d) Prediction for original image (e) Target (f) Prediction for perturbed image

Figure 18: Targeted image-dependent perturbations with $L_\infty = 10$.

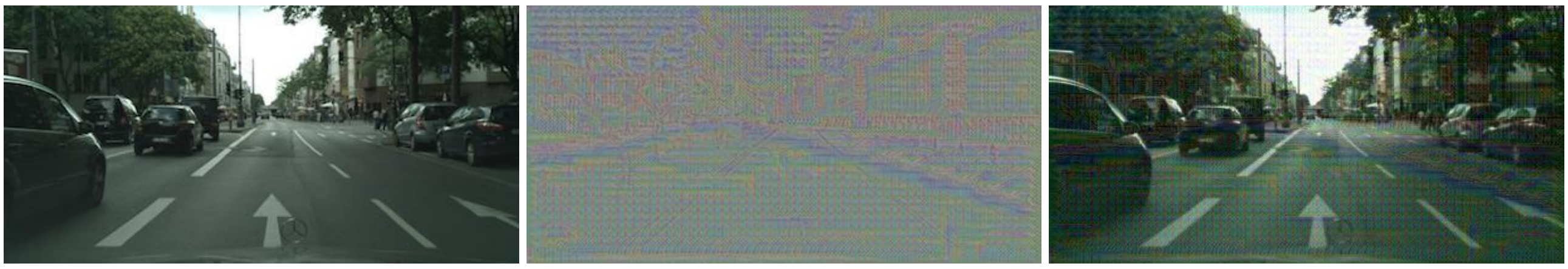

(a) Original image (b) Perturbation (c) Perturbed image

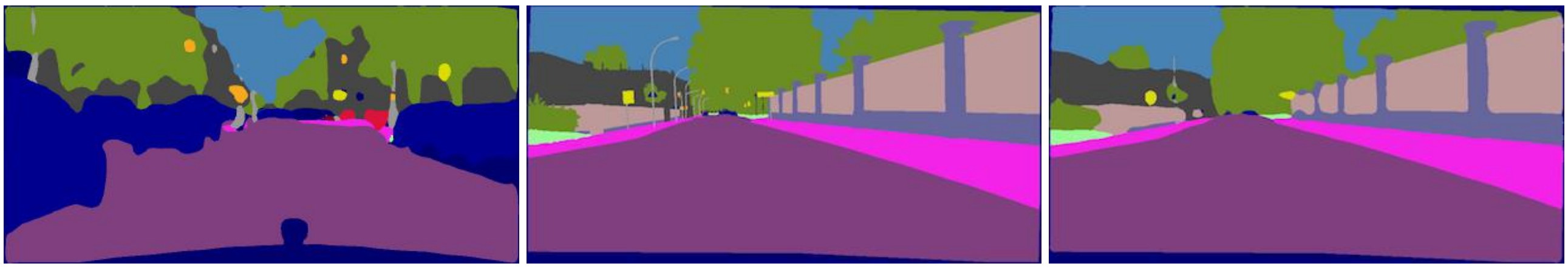

(d) Prediction for original image (e) Target (f) Prediction for perturbed image

Figure 19: Targeted image-dependent perturbations with $L_\infty = 20$.

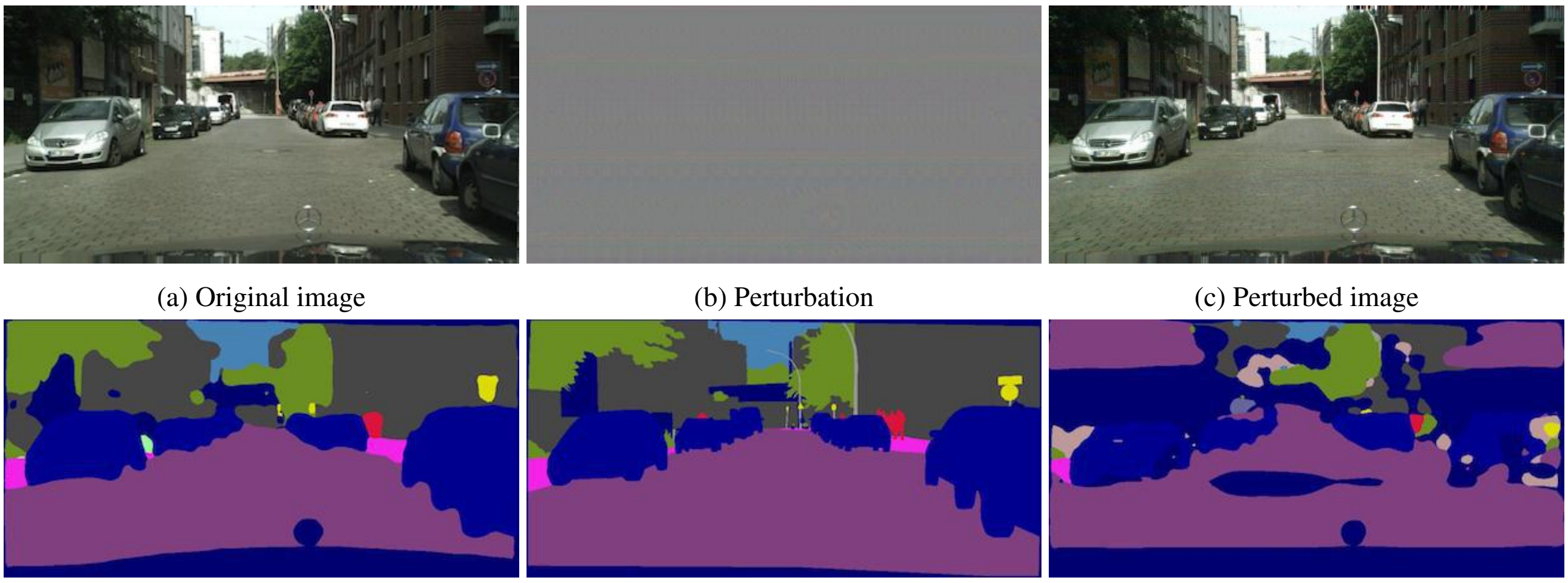

(a) Original image (b) Perturbation (c) Perturbed image

(d) Prediction for original image (e) Groundtruth (f) Prediction for perturbed image

Figure 20: Non-targeted universal perturbations with $L_\infty = 5$. Zoom in for details.

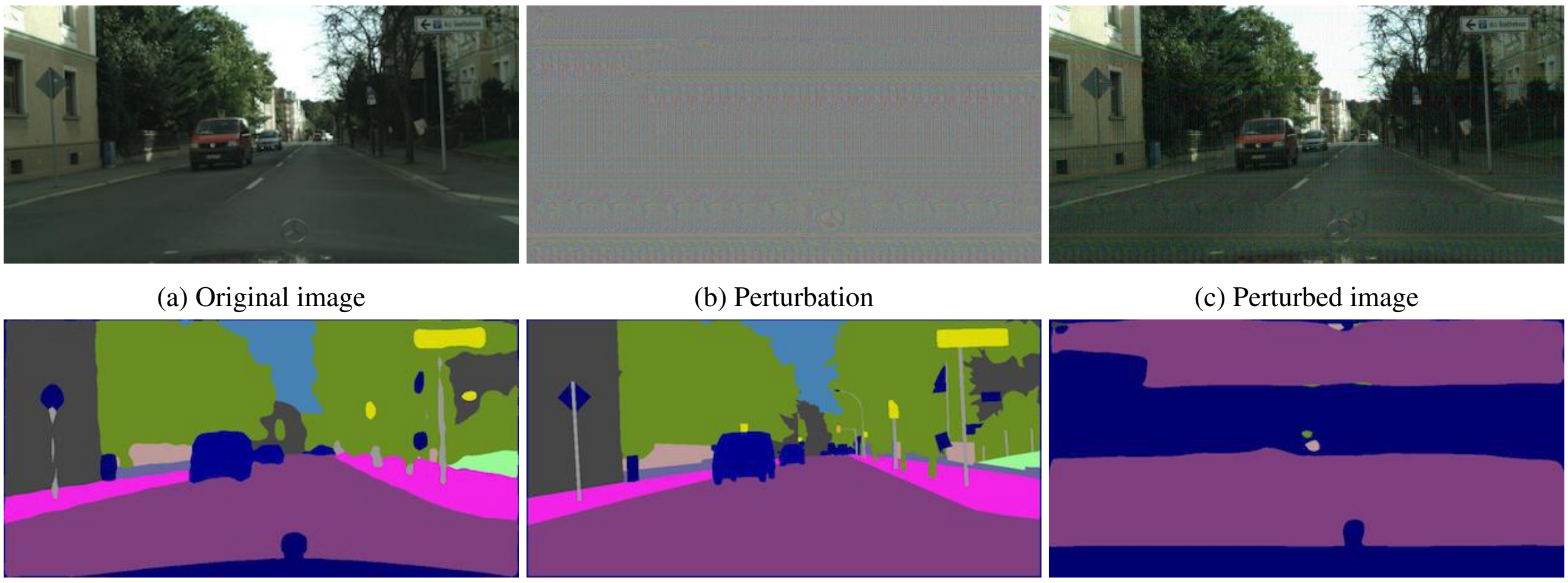

(a) Original image (b) Perturbation (c) Perturbed image

(d) Prediction for original image (e) Groundtruth (f) Prediction for perturbed image

Figure 21: Non-targeted universal perturbations with $L_\infty = 10$.

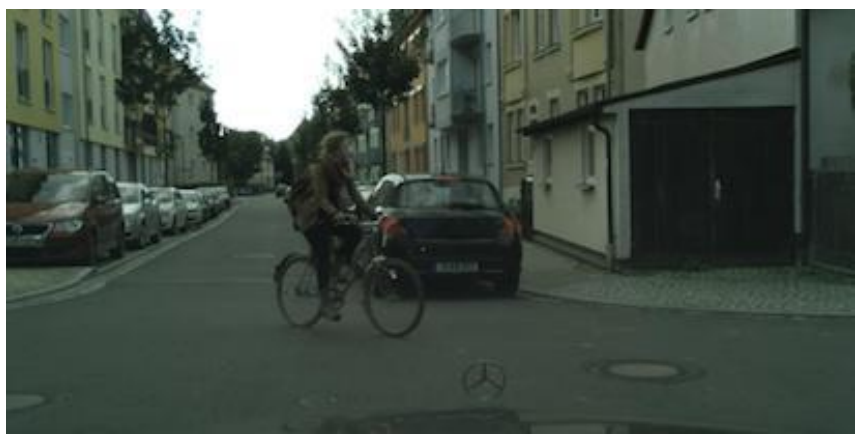
(a) Original image

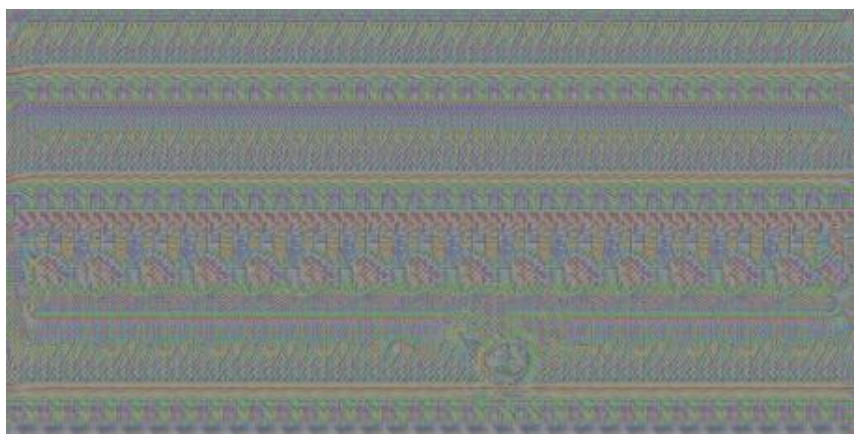
(b) Perturbation

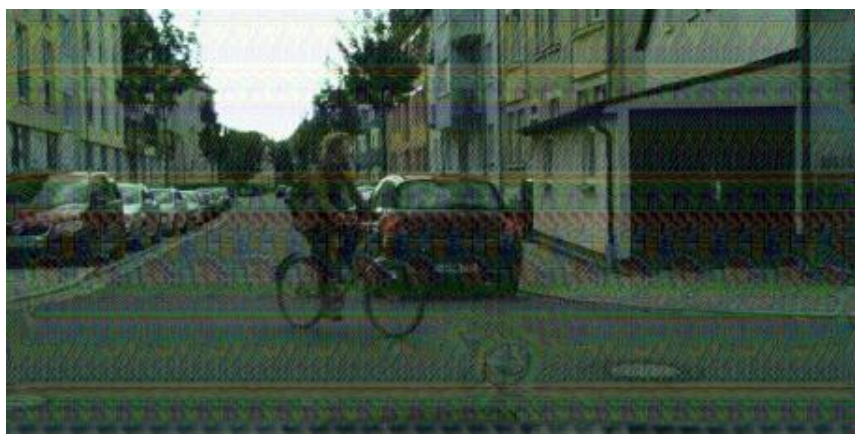
(c) Perturbed image

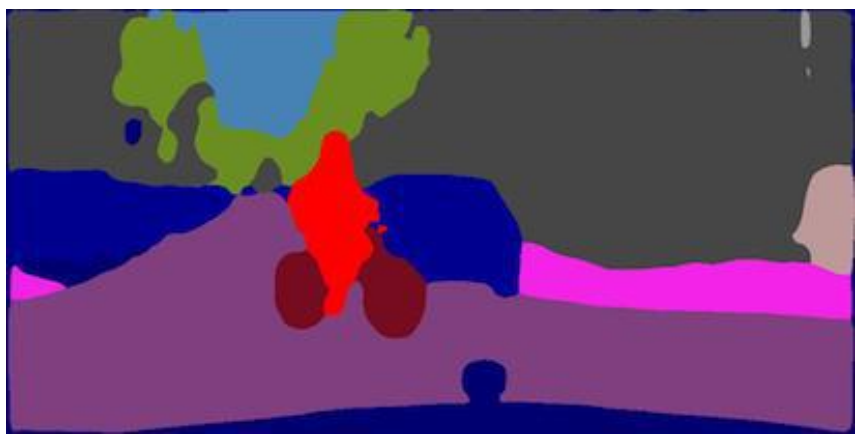
(d) Prediction for original image

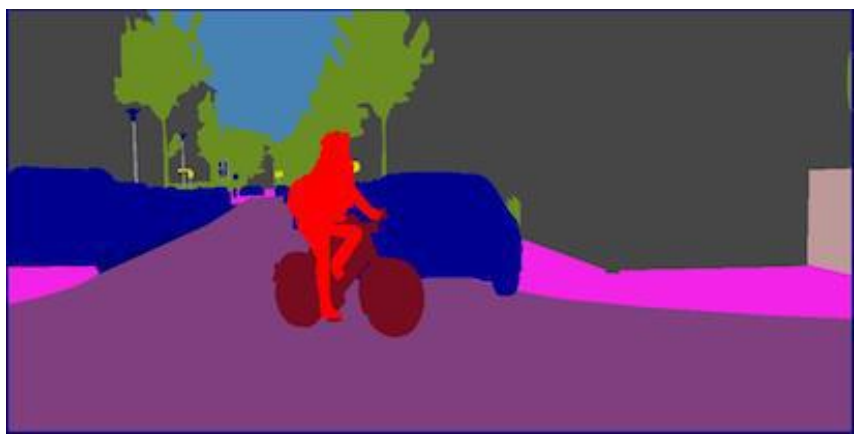
(e) Groundtruth

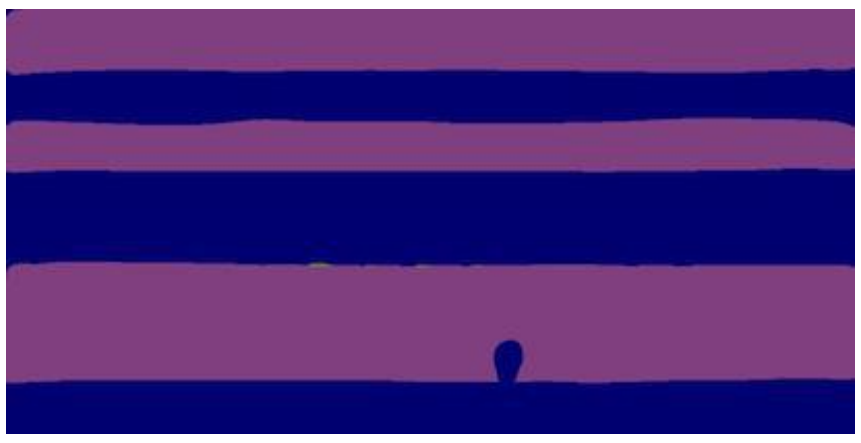
(f) Prediction for perturbed image

Figure 22: Non-targeted universal perturbations with $L_\infty = 20$.

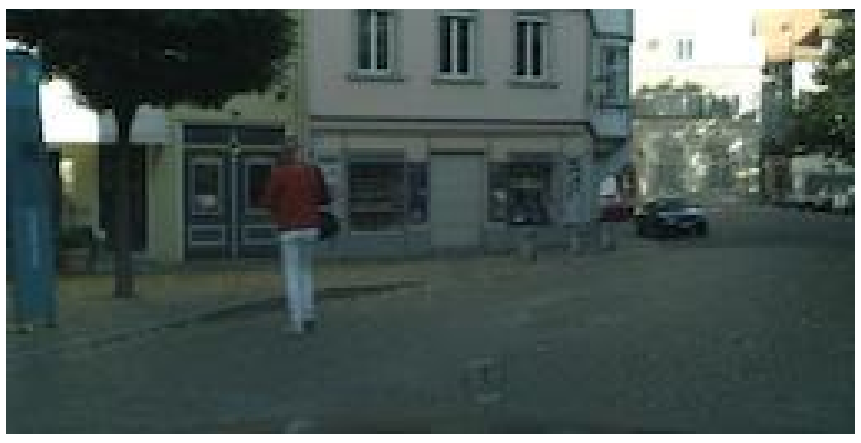
(a) Original image

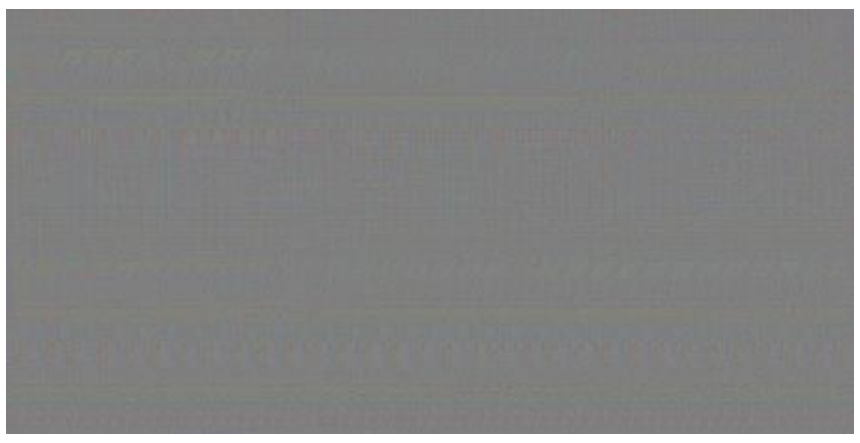
(b) Perturbation

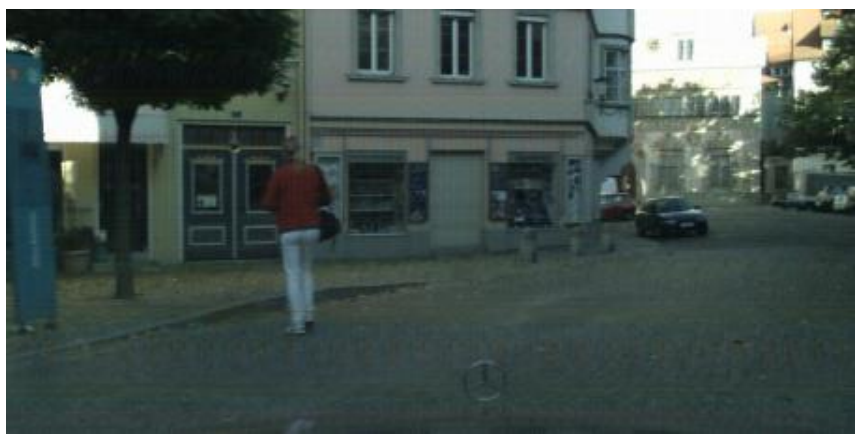
(c) Perturbed image

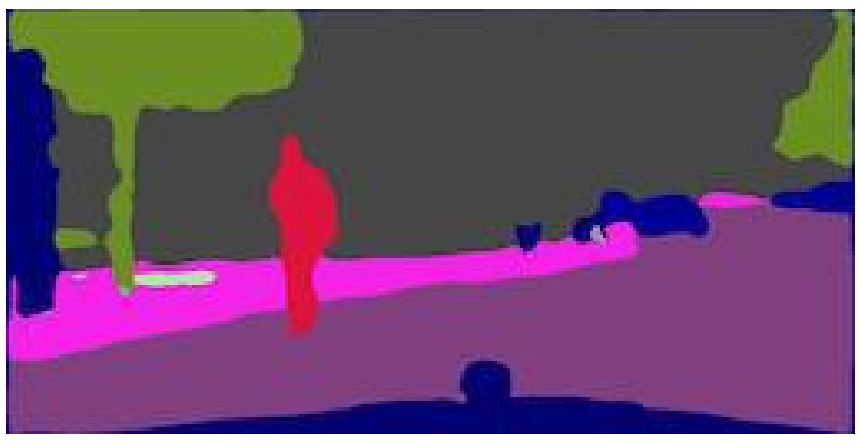
(d) Prediction for original image

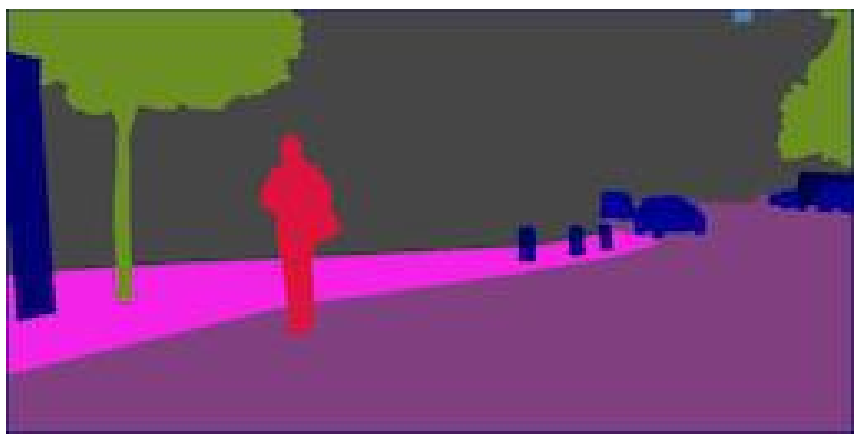
(e) Groundtruth

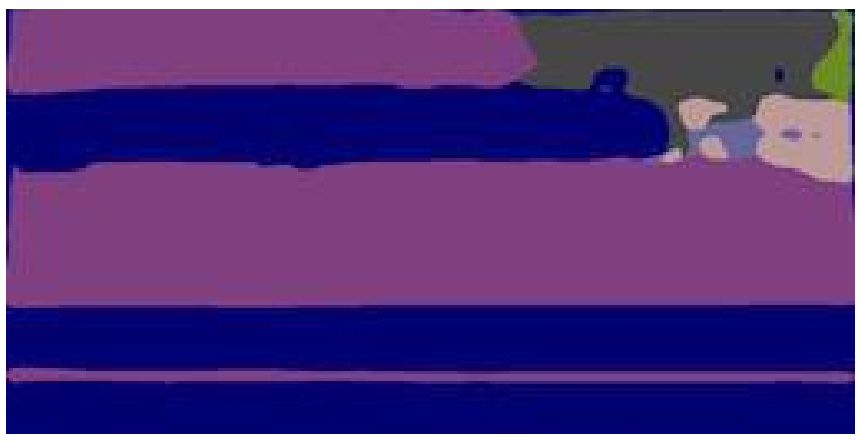
(f) Prediction for perturbed image

Figure 23: Non-targeted image-dependent perturbations with $L_\infty = 5$. Zoom in for details.

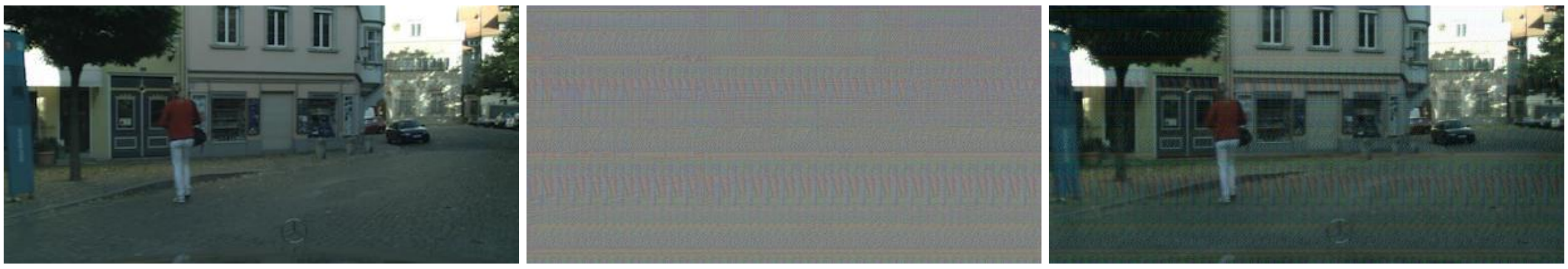

(a) Original image (b) Perturbation (c) Perturbed image

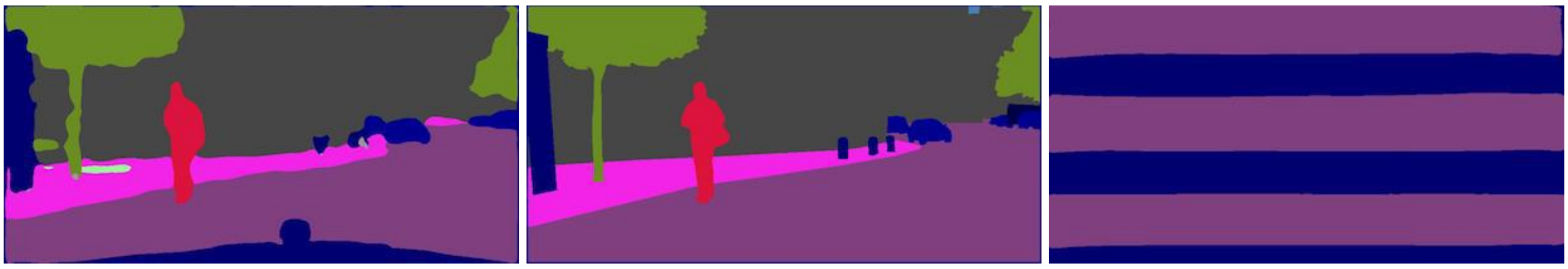

(d) Prediction for original image (e) Groundtruth (f) Prediction for perturbed image

Figure 24: Non-targeted image-dependent perturbations with $L_\infty = 10$.

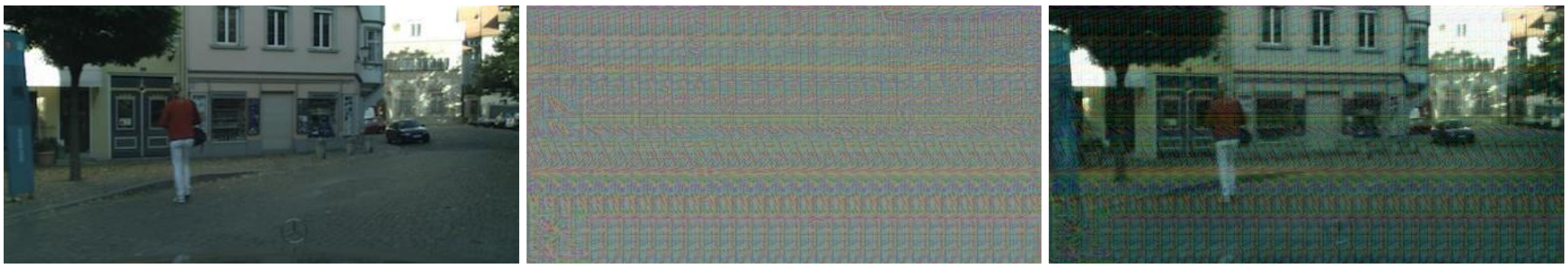

(a) Original image (b) Perturbation (c) Perturbed image

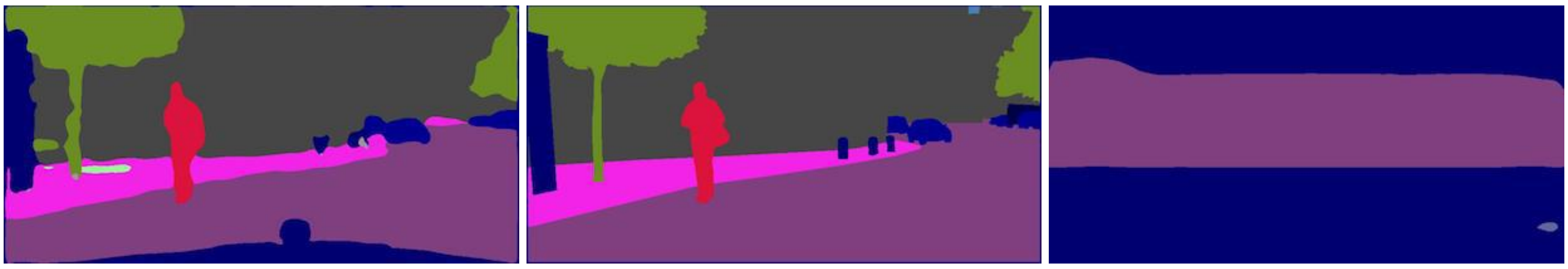

(d) Prediction for original image (e) Groundtruth (f) Prediction for perturbed image

Figure 25: Non-targeted image-dependent perturbations with $L_\infty = 20$.

## References

[1] N. Akhtar, J. Liu, and A. Mian. Defense against universal adversarial perturbations. *arXiv preprint arXiv:1711.05929*, 2017. 2

[2] A. Arnab, O. Miksik, and P. H. Torr. On the robustness of semantic segmentation models to adversarial attacks. *arXiv preprint arXiv:1711.09856*, 2017. 2

[3] A. Athalye, N. Carlini, and D. Wagner. Obfuscated gradients give a false sense of security: Circumventing defenses to adversarial examples. *arXiv preprint arXiv:1802.00420*, 2018. 2

[4] A. Athalye and I. Sutskever. Synthesizing robust adversarial examples. *arXiv preprint arXiv:1707.07397*, 2017. 2

[5] V. Badrinarayanan, A. Kendall, and R. Cipolla. Segnet: A deep convolutional encoder-decoder architecture for image segmentation. *arXiv preprint arXiv:1511.00561*, 2015. 1

[6] S. Baluja and I. Fischer. Adversarial transformation networks: Learning to generate adversarial examples. *arXiv preprint arXiv:1703.09387*, 2017. 2

[7] A. N. Bhagoji, W. He, B. Li, and D. Song. Exploring the space of black-box attacks on deep neural networks. *arXiv preprint arXiv:1712.09491*, 2017. 3, 5, 6

[8] N. Carlini and D. Wagner. Towards evaluating the robustness of neural networks. In *Security and Privacy (SP), 2017 IEEE Symposium on*, pages 39–57. IEEE, 2017. 2, 3, 5, 6

[9] L.-C. Chen, G. Papandreou, I. Kokkinos, K. Murphy, and A. L. Yuille. Deeplab: Semantic image segmentation with deep convolutional nets, atrous convolution, and fully connected crfs. *arXiv preprint arXiv:1606.00915*, 2016. 1

[10] P.-Y. Chen, H. Zhang, Y. Sharma, J. Yi, and C.-J. Hsieh. Zoo: Zeroth order optimization based black-box attacks to deep neural networks without training substitute models. In *Proceedings of the 10th ACM Workshop on Artificial Intelligence and Security*, pages 15–26. ACM, 2017. 6

[11] M. Cisse, Y. Adi, N. Neverova, and J. Keshet. Houdini: Fooling deep structured prediction models. *arXiv preprint arXiv:1707.05373*, 2017. 2

[12] M. Cordts, M. Omran, S. Ramos, T. Rehfeld, M. Enzweiler, R. Benenson, U. Franke, S. Roth, and B. Schiele. The cityscapes dataset for semantic urban scene understanding. In *Proceedings of the IEEE Conference on Computer Vision and Pattern Recognition*, pages 3213–3223, 2016. 7

[13] N. Das, M. Shanbhogue, S.-T. Chen, F. Hohman, S. Li, L. Chen, M. E. Kounavis, and D. H. Chau. Shield: Fast, practical defense and vaccination for deep learning using jpeg compression. *arXiv preprint arXiv:1802.06816*, 2018. 2

[14] J. Deng, W. Dong, R. Socher, L.-J. Li, K. Li, and L. Fei-Fei. Imagenet: A large-scale hierarchical image database. In *Computer Vision and Pattern Recognition, 2009. CVPR 2009. IEEE Conference on*, pages 248–255. IEEE, 2009. 5

[15] E. L. Denton, S. Chintala, R. Fergus, et al. Deep generative image models using a laplacian pyramid of adversarial networks. In *Advances in neural information processing systems*, pages 1486–1494, 2015. 3

[16] G. S. Dhillon, K. Azizzadenesheli, Z. C. Lipton, J. Bernstein, J. Kossaifi, A. Khanna, and A. Anandkumar. Stochastic activation pruning for robust adversarial defense. *arXiv preprint arXiv:1803.01442*, 2018. 2

[17] I. Goodfellow, J. Pouget-Abadie, M. Mirza, B. Xu, D. Warde-Farley, S. Ozair, A. Courville, and Y. Bengio. Generative adversarial nets. In *Advances in neural information processing systems*, pages 2672–2680, 2014. 3

[18] I. J. Goodfellow, J. Shlens, and C. Szegedy. Explaining and harnessing adversarial examples. *arXiv preprint arXiv:1412.6572*, 2014. 2, 6

[19] C. Guo, M. Rana, M. Cissé, and L. van der Maaten. Countering adversarial images using input transformations. *arXiv preprint arXiv:1711.00117*, 2017. 2

[20] K. He, X. Zhang, S. Ren, and J. Sun. Deep residual learning for image recognition. In *Proceedings of the IEEE conference on computer vision and pattern recognition*, pages 770–778, 2016. 1

[21] X. Huang, Y. Li, O. Poursaeed, J. Hopcroft, and S. Belongie. Stacked generative adversarial networks. *arXiv preprint arXiv:1612.04357*, 2016. 3

[22] P. Isola, J.-Y. Zhu, T. Zhou, and A. A. Efros. Image-to-image translation with conditional adversarial networks. *arXiv preprint arXiv:1611.07004*, 2016. 3

[23] J. Johnson, A. Alahi, and L. Fei-Fei. Perceptual losses for real-time style transfer and super-resolution. In *European Conference on Computer Vision*, pages 694–711. Springer, 2016. 3

[24] D. Kingma and J. Ba. Adam: A method for stochastic optimization. *arXiv preprint arXiv:1412.6980*, 2014. 2

[25] A. Krizhevsky, I. Sutskever, and G. E. Hinton. Imagenet classification with deep convolutional neural networks. In *Advances in neural information processing systems*, pages 1097–1105, 2012. 1

[26] A. Kurakin, I. Goodfellow, and S. Bengio. Adversarial examples in the physical world. *arXiv preprint arXiv:1607.02533*, 2016. 2, 3, 24

[27] A. Kurakin, I. Goodfellow, and S. Bengio. Adversarial machine learning at scale. *arXiv preprint arXiv:1611.01236*, 2016. 2, 3

[28] A. B. L. Larsen, S. K. Sønderby, H. Larochelle, and O. Winther. Autoencoding beyond pixels using a learned similarity metric. *arXiv preprint arXiv:1512.09300*, 2015. 3

[29] Y. Liu, X. Chen, C. Liu, and D. Song. Delving into transferable adversarial examples and black-box attacks. *arXiv preprint arXiv:1611.02770*, 2016. 6

[30] J. Long, E. Shelhamer, and T. Darrell. Fully convolutional networks for semantic segmentation. In *Proceedings of the IEEE Conference on Computer Vision and Pattern Recognition*, pages 3431–3440, 2015. 1, 7

[31] J. Lu, H. Sibai, E. Fabry, and D. Forsyth. No need to worry about adversarial examples in object detection in autonomous vehicles. *arXiv preprint arXiv:1707.03501*, 2017. 2

[32] X. Ma, B. Li, Y. Wang, S. M. Erfani, S. Wijewickrema, M. E. Houle, G. Schoenebeck, D. Song, and J. Bailey. Characterizing adversarial subspaces using local intrinsic dimensionality. *arXiv preprint arXiv:1801.02613*, 2018. 2

[33] A. Madry, A. Makelov, L. Schmidt, D. Tsipras, and A. Vladu. Towards deep learning models resistant to adversarial attacks. *arXiv preprint arXiv:1706.06083*, 2017. 2

[34] J. H. Metzen, M. C. Kumar, T. Brox, and V. Fischer. Universal adversarial perturbations against semantic image segmentation. *arXiv preprint arXiv:1704.05712*, 2017. 2, 6, 7, 8

[35] S.-M. Moosavi-Dezfooli, A. Fawzi, O. Fawzi, and P. Frossard. Universal adversarial perturbations. *arXiv preprint arXiv:1610.08401*, 2016. 1, 2, 5

[36] S.-M. Moosavi-Dezfooli, A. Fawzi, O. Fawzi, P. Frossard, and S. Soatto. Analysis of universal adversarial perturbations. *arXiv preprint arXiv:1705.09554*, 2017. 2

[37] S.-M. Moosavi-Dezfooli, A. Fawzi, and P. Frossard. Deepfool: a simple and accurate method to fool deep neural networks. In *Proceedings of the IEEE Conference on Computer Vision and Pattern Recognition*, pages 2574–2582, 2016. 2

[38] K. R. Mopuri, U. Garg, and R. V. Babu. Fast feature fool: A data independent approach to universal adversarial perturbations. *arXiv preprint arXiv:1707.05572*, 2017. 2

[39] A. Nguyen, J. Yosinski, and J. Clune. Deep neural networks are easily fooled: High confidence predictions for unrecognizable images. In *Proceedings of the IEEE Conference on Computer Vision and Pattern Recognition*, pages 427–436, 2015. 2

[40] N. Papernot, P. McDaniel, and I. Goodfellow. Transferability in machine learning: from phenomena to black-box attacks using adversarial samples. *arXiv preprint arXiv:1605.07277*, 2016. 6

[41] N. Papernot, P. McDaniel, I. Goodfellow, S. Jha, Z. B. Celik, and A. Swami. Practical black-box attacks against deep learning systems using adversarial examples. *arXiv preprint arXiv:1602.02697*, 2016. 6

[42] A. Prakash, N. Moran, S. Garber, A. DiLillo, and J. Storer. Deflecting adversarial attacks with pixel deflection. *arXiv preprint arXiv:1801.08926*, 2018. 2

[43] A. Radford, L. Metz, and S. Chintala. Unsupervised representation learning with deep convolutional generative adversarial networks. *arXiv preprint arXiv:1511.06434*, 2015. 3

[44] A. Raghunathan, J. Steinhardt, and P. Liang. Certified defenses against adversarial examples. *arXiv preprint arXiv:1801.09344*, 2018. 2

[45] A. S. Rakin, Z. He, B. Gong, and D. Fan. Robust preprocessing: A robust defense method against adversary attack. *arXiv preprint arXiv:1802.01549*, 2018. 2

[46] O. Ronneberger, P. Fischer, and T. Brox. U-net: Convolutional networks for biomedical image segmentation. In *International Conference on Medical Image Computing and Computer-Assisted Intervention*, pages 234–241. Springer, 2015. 3

[47] A. Roy, C. Raffel, I. Goodfellow, and J. Buckman. Thermometer encoding: One hot way to resist adversarial examples. 2018. 2

[48] P. Samangouei, M. Kabkab, and R. Chellappa. Defense-gan: Protecting classifiers against adversarial attacks using generative models. 2018. 2

[49] K. Simonyan and A. Zisserman. Very deep convolutional networks for large-scale image recognition. *arXiv preprint arXiv:1409.1556*, 2014. 1

[50] Y. Song, T. Kim, S. Nowozin, S. Ermon, and N. Kushman. Pixeldefend: Leveraging generative models to understand and defend against adversarial examples. *arXiv preprint arXiv:1710.10766*, 2017. 2

[51] C. Szegedy, W. Liu, Y. Jia, P. Sermanet, S. Reed, D. Anguelov, D. Erhan, V. Vanhoucke, and A. Rabinovich. Going deeper with convolutions. In *Proceedings of the IEEE conference on computer vision and pattern recognition*, pages 1–9, 2015. 1

[52] C. Szegedy, V. Vanhoucke, S. Ioffe, J. Shlens, and Z. Wojna. Rethinking the inception architecture for computer vision. In *Proceedings of the IEEE Conference on Computer Vision and Pattern Recognition*, pages 2818–2826, 2016. 1

[53] C. Szegedy, W. Zaremba, I. Sutskever, J. Bruna, D. Erhan, I. Goodfellow, and R. Fergus. Intriguing properties of neural networks. *arXiv preprint arXiv:1312.6199*, 2013. 1, 2, 6

[54] F. Tramèr, A. Kurakin, N. Papernot, D. Boneh, and P. McDaniel. Ensemble adversarial training: Attacks and defenses. *arXiv preprint arXiv:1705.07204*, 2017. 2

[55] D. Vijaykeerthy, A. Suri, S. Mehta, and P. Kumaraguru. Hardening deep neural networks via adversarial model cascades. *arXiv preprint arXiv:1802.01448*, 2018. 2

[56] T.-W. Weng, H. Zhang, P.-Y. Chen, J. Yi, D. Su, Y. Gao, C.-J. Hsieh, and L. Daniel. Evaluating the robustness of neural networks: An extreme value theory approach. *arXiv preprint arXiv:1801.10578*, 2018. 2

[57] C. Xie, J. Wang, Z. Zhang, Z. Ren, and A. Yuille. Mitigating adversarial effects through randomization. *arXiv preprint arXiv:1711.01991*, 2017. 2

[58] C. Xie, J. Wang, Z. Zhang, Y. Zhou, L. Xie, and A. Yuille. Adversarial examples for semantic segmentation and object detection. *arXiv preprint arXiv:1703.08603*, 2017. 2, 6

[59] F. Yu and V. Koltun. Multi-scale context aggregation by dilated convolutions. *arXiv preprint arXiv:1511.07122*, 2015. 1

[60] H. Zhao, J. Shi, X. Qi, X. Wang, and J. Jia. Pyramid scene parsing network. *arXiv preprint arXiv:1612.01105*, 2016. 1

[61] J.-Y. Zhu, T. Park, P. Isola, and A. A. Efros. Unpaired image-to-image translation using cycle-consistent adversarial networks. *arXiv preprint arXiv:1703.10593*, 2017. 3

# 7. Appendix

## 7.1. Runtime Analysis

Note that inference time is not an issue for universal perturbations as we just need to add the perturbation to the input image during inference. Therefore, we provide running time only for image-dependent perturbations. In this case, we need to forward the input image to the generator and get the resulting perturbation. Table 8 demonstrates the inference time for image-dependent perturbations. It also shows the generator's architecture for each task including the number of filters in the first layer. We perform model-level parallelization across two GPUs, and batch size is set to be one. Notice that inference time is in the order of milliseconds, allowing us to generate perturbations in real-time. Table 9 shows inference time for the segmentation task. Two architectures with similar performance are given. Here we deal with $1024 \times 512$ images in the Cityscapes dataset, and we need models with more capacity; hence, the inference time is larger compared with the classification task.

| Task | Architecture | Titan Xp | Tesla K40 |
|---|---|---|---|
| Non-targeted | ResNet Gen. 6 blocks, 50 filters | 0.27 ms | 4.7 ms |
| Targeted | ResNet Gen. 6 blocks, 57 filters | 0.28 ms | 4.8 ms |

Table 8: Average inference time per image and generator's architecture for image-dependent classification tasks. Target model is Inception-v3.

| Architecture | Titan Xp | Tesla K40m |
|---|---|---|
| U-Net Generator: 8 layers, 200 filters | 132.8 ms | 511.7 ms |
| ResNet Generator: 9 blocks, 145 filters | 335.7 ms | 2396.9 ms |

Table 9: Average inference time per image and generator's architecture for the semantic segmentation task. Targeted image-dependent perturbations are considered with FCN-8s as the pre-trained model.

## 7.2. Resistance to Gaussian Blur

We examine the effect of applying Gaussian filters to perturbed images. Results for the classification task are shown in Table 10. In order to be comparable with [26], we consider non-targeted image-dependent perturbations with Destruction Rate (fraction of images that are no longer misclassified after blur) as the metric. For most $\sigma$ values, our method is more resistant to Gaussian blur than I-FGSM.

We also evaluate the effect of Gaussian filters for the segmentation task. Results are given in Table 11. As we can observe, the perturbations are reasonably robust to Gaussian blur.

| | $\sigma = 0.5$ | $\sigma = 0.75$ | $\sigma = 1$ | $\sigma = 1.25$ |
|---|---|---|---|---|
| GAP | 0.0% | 0.8% | 3.2% | 8.0% |
| I-FGSM | 0.0% | 0.5% | 8.0% | 23.0% |

Table 10: Destruction Rate of non-targeted image-dependent perturbations for the classification task. Perturbation norm is set to $L_\infty = 16$.

| | $\sigma = 0.5$ | $\sigma = 0.75$ | $\sigma = 1$ | $\sigma = 1.25$ |
|---|---|---|---|---|
| $L_\infty = 5$ | 83.2% | 76.9% | 66.0% | 57.1 % |
| $L_\infty = 10$ | 94.8% | 90.1% | 80.0% | 69.6% |
| $L_\infty = 20$ | 97.5% | 95.7% | 89.3% | 78.8% |

Table 11: Success rate of targeted image-dependent perturbations for the segmentation task after applying Gaussian filters.